\pdfoutput=1
\documentclass[fullname]{clv2}
\makeatletter
\newcommand{\@BIBLABEL}{\@emptybiblabel}
\newcommand{\@emptybiblabel}[1]{}
\makeatother
\usepackage{url}
\usepackage{hyperref}

\issue{X}{Y}{2011}

\dochead{Modality and Negation in SIMT}

\runningtitle{Modality and Negation in SIMT}

\runningauthor{Baker et al.}
\historydates{Submission received: 27 March 2011; revised submission received: 28 September 2011; accepted for publication: 30 November 2011. 
Manuscript was published in Computational Linguistics 38(2):411-438. 
Journal website: http://www.mitpressjournals.org/loi/coli}

\begin{document}

\title{Use of Modality and Negation in Semantically-Informed Syntactic MT}

\author{Kathryn Baker}
\affil{U.S. Dept. of Defense\thanks{Fort Meade, MD, klbake4@tycho.ncsc.mil}}

\author{Michael Bloodgood and\\ Bonnie J. Dorr}
\affil{University of Maryland\thanks{Institute for Advanced Computer Studies and Center for Advanced Study of Language, College Park, MD, bonnie@umiacs.umd.edu, meb@umd.edu}}

\author{Chris Callison-Burch and Nathaniel W. Filardo and Christine Piatko}
\affil{Johns Hopkins University\thanks{Human Language Technology Center of Excellence, Baltimore, MD and Applied Physics Laboratory, Laurel, MD, \{ccb,nwf\}@cs.jhu.edu, christine.piatko@jhuapl.edu}}

\author{Lori Levin}
\affil{Carnegie Mellon University\thanks{Pittsburgh, PA, lsl@cs.cmu.edu}}

\author{Scott Miller}
\affil{BBN Technologies\thanks{Cambridge, MA, smiller@bbn.edu}}

\maketitle

\begin{abstract}
This paper describes the resource- and system-building efforts of an
eight-week Johns Hopkins University Human Language Technology Center
of Excellence {\it Summer Camp for Applied Language Exploration\/}
(SCALE-2009) on Semantically-Informed Machine Translation (SIMT). We
describe a new modality/negation (MN) annotation scheme, the creation
of a (publicly available) MN lexicon, and two automated MN taggers
that we built using the annotation scheme and lexicon.  Our annotation
scheme isolates three components of modality and negation: a trigger
(a word that conveys modality or negation), a target (an action
associated with modality or negation) and a holder (an experiencer of
modality). We describe how our MN lexicon was semi-automatically
produced and we demonstrate that a structure-based MN tagger results
in precision around 86\% (depending on genre) for tagging of a
standard LDC data set.

We apply our MN annotation scheme to statistical machine translation using
a syntactic framework that
supports the inclusion of semantic annotations.  Syntactic tags enriched with
semantic annotations are assigned to parse trees in the target-language 
training texts through a process of tree grafting.  While the focus of our work is
modality and negation, the tree grafting procedure is general and supports
other types of semantic information.  We exploit this capability by including
named entities, produced by a pre-existing tagger, in addition to the MN elements
produced by the taggers described in this paper. The resulting
system significantly outperformed a
linguistically na\"{\i}ve baseline model (Hiero), and reached the highest
scores yet reported on the NIST 2009 Urdu-English test set.  This
finding supports the hypothesis that both syntactic and semantic
information can improve translation quality.
\end{abstract}

\section{Introduction}

This paper describes the resource- and system-building efforts of an
eight-week Johns Hopkins Human Language Technology Center of Excellence {\it
  Summer Camp for Applied Language Exploration\/} (SCALE-2009) on
Semantically-Informed Machine Translation (SIMT)
\cite{Baker:2010a,Baker:2010d,Baker:2010b,Baker:2010c}.  Specifically,
we describe our modality/negation (MN) annotation scheme, a
(publicly available) MN lexicon, and two automated MN taggers that
were built using the lexicon and annotation scheme.  

Our annotation scheme isolates three components of modality and
negation: a trigger (a word that conveys modality or negation), a
target (an action associated with modality or negation) and a holder
(an experiencer of modality). Two examples of MN tagging are shown in
Figure~\ref{modality-example}.  

\begin{figure}[h]
\begin{center}
\begin{small}
\begin{tabular}{lp{5in}}
(1)&{\bf Input:} Americans should know that we can not hand over Dr. Khan to them.\\
   &{\bf Output:} Americans \verb|<|TrigRequire should\verb|>|  \verb|<|TargRequire know\verb|>| that we \verb|<|TrigAble can\verb|>|  \verb|<|TrigNegation not\verb|>| \verb|<|TargNOTAble hand\verb|>|  over Dr. Khan to them.\\  \\
(2)&{\bf Input:} He managed to hold general elections in the year 2002, but he can not be ignorant of the fact that the world at large did not accept these elections.\\
   &{\bf Output:} He \verb|<|TrigSucceed managed\verb|>| to \verb|<|TargSucceed hold\verb|>| general elections in the year 2002, but he \verb|<|TrigAble can\verb|>|  \verb|<|TrigNegation not\verb|>| \verb|<|TargNOTAble be\verb|>| ignorant of the fact that the world at large did \verb|<|TrigNegation not\verb|>| \verb|<|TrigBelief accept\verb|>| these \verb|<|TargNOTBelief elections\verb|>|.
\end{tabular}
\end{small}
\end{center}
\caption{Modality/Negation Tagging Examples.}
\label{modality-example}
\end{figure}

Note that modality and negation are
unified into single MN tags (e.g., the ``Able'' modality tag is
combined with ``NOT'' to form the ``NOTAble'' tag) and also that MN
tags occur in pairs of triggers (e.g., TrigAble and TrigNegation) and
targets (e.g., TargNOTAble).

We apply our modality and negation mechanism to the problem of 
Urdu-English machine translation using a technique that we call 
{\it tree grafting\/}. This technique incorporates syntactic labels 
and semantic annotations in a unified and coherent framework for implementing 
semantically-informed machine translation.  Our framework is not limited to the
semantic annotations produced by the MN taggers that are the subject of this paper and
 we exploit this capability to additionally include named-entity annotations produced 
 by a pre-existing tagger.  By augmenting hierarchical phrase-based
translation rules with syntactic labels that were extracted from a
parsed parallel corpus, and further augmenting the parse trees with
markers for modality, negation, and entities (through the tree grafting 
process), we produced a better model for
translating Urdu and English. The resulting system significantly
outperformed the linguistically na\"{\i}ve baseline Hiero model, and
reached the highest scores yet reported on the NIST 2009 Urdu-English
translation task.

We note that while our largest gains were from syntactic enrichments
to the model, smaller (but significant) gains were achieved by
injecting semantic knowledge into the syntactic paradigm.  Verbal
semantics (modality and negation) contributed slightly more gains than
nominal semantics (named entities) while their combined gains were
the sum of their individual contributions.

Of course, the limited semantic types we explored (modality, negation and entities) 
are only a small piece of
the much larger semantic space, but demonstrating success on these
semantic aspects of language, the combination of which 
has been unexplored by the 
statistical machine translation community, bodes well for (larger)
improvements based on the incorporation of other semantic aspects
(e.g., relations and temporal knowledge).  Moreover, we believe this
syntactic framework to be well suited for further exploration of the
impact of many different types of semantics on the quality
of machine-translation (MT) output.  Indeed, it
would not have been possible to initiate the current study without the
foundational work that gave rise to a syntactic paradigm that could
support these semantic enrichments.

In the SIMT paradigm, semantic elements (e.g., modality/negation) are
identified in the English portion of a parallel training corpus and
projected to the source language (in our case, Urdu) during a process
of syntactic alignment.  These semantic elements are subsequently used in the translation rules that are extracted from the parallel corpus.  The goal of adding them to the translation rules is to constrain the space of possible
translations to more grammatical and more semantically coherent output.  We explored whether including such semantic elements
could improve translation output in the face
of sparse training data and few source language annotations.  Results
were encouraging.  Translation quality, as measured by the Bleu metric
\cite{Papineni:2002}, improved when the training process for the
Joshua machine translation system \cite{Li:2009} used in the SCALE workshop
included MN annotation.

We were particularly interested in identifying modalities and negation
because they can be used to characterize events in a variety of
automated analytic processes.  Modalities and negation can distinguish realized
events from unrealized events, beliefs from certainties, and can
distinguish positive and negative instances of entities and events.
For example, the correct identification and retention of negation in a
particular language---such as a single instance of the word
``not''---is very important for a correct representation of events and
likewise for translation.

The next two sections examine related work and the motivation behind
the SIMT approach.  Section~\ref{modality-section} defines the
theoretical framework for our MN lexicon and automatic MN taggers.
Section~\ref{modality-annotation-scheme} presents the MN annotation
scheme used by our human annotators and describes the creation of a MN
lexicon based on this scheme.
Section~\ref{Automatic-Modality-Annotation} presents two types of MN
taggers---one that is string-based and one that is
structure-based---and evaluates the effectiveness of the
structure-based tagger.  Section~\ref{joshua} then presents
implementation details of the semantically-informed syntactic system
and describes the results of its application.  Finally,
Section~\ref{conclusions} presents conclusions and future work.

\section{Related Work}
\label{related-work}

The development of annotation schemes has become an area of
computational linguistics development in its own right, often separate
from machine learning applications.  Some projects began as strictly
linguistic projects that were later adapted for computational
linguistics.  When an annotation scheme is consistent and well
developed, its subsequent application to NLP systems is most
effective. For example, the syntactic annotation of parse trees in the
Penn Treebank~\cite{Marcus:93} had a tremendous effect on parsing and
on Natural Language Processing in general.

In the case of semantic annotations, each tends to have its unique
area of focus.  While the labeling conventions may differ, a layer of
modality annotation over verb role annotation, for example, can have a
complementary effect of providing more information, rather than being
viewed as a competing scheme.  We review some of the major semantic
annotation efforts below.

Propbank~\cite{Palmer:2005} is a set of annotations of
predicate-argument structure over parse trees.  First annotated as an
overlay to the Penn Treebank, Propbank annotation now exists for other
corpora.  Propbank annotation aims to answer the question {\it Who did
 what to whom?\/} for individual predicates.  It is tightly coupled
with the behavior of individual verbs.  FrameNet~\cite{Baker:1998}, a
frame-based lexical database that associates each word in the database
with a semantic frame and semantic roles, is also associated with
annotations at the lexical level.  WordNet~\cite{fellbaum98wordnet} is
a very widely used online lexical taxonomy which has been developed in
numerous languages.  WordNet nouns, verbs, adjectives and adverbs are
organized into synonym sets.  PropBank, FrameNet, and WordNet cover
the word senses and argument-taking properties of many modal
predicates.

The Prague Dependency Treebank~\cite{PDT10,trmanEn2005} (PDT) is a
multi-level system of annotation for texts in Czech and other
languages, with its roots in the Prague school of linguistics.
Besides a morphological layer and an analytical layer, there is a
Tectogrammatical layer.  The Tectogrammatical layer includes
functional relationships, dependency relations and co-reference.  The
PDT also integrates propositional and extra-propositional meanings in
a single annotation framework.

The Penn Discourse Treebank (PDTB)~\cite{Webber:2003,PRASAD08.754}
annotates discourse connectives and their arguments over a portion of
the Penn Treebank.  Within this framework, senses are annotated for
the discourse connectives in a hierarchical scheme.  Relevant to the
current work, one type of tag in the scheme is the Conditional tag,
which includes hypothetical, general, unreal present, unreal past, factual
present, and factual past arguments.

The PDTB work is related to that of Wiebe, Wilson, and Cardie~\shortcite{Wiebe05} for establishing
the importance of attributing a belief or assertion expressed in text
to its agent (equivalent to the notion of {\it holder\/} in our
scheme).  The annotation scheme is designed to capture the
expression of opinions and emotions.  In the PDTB, each discourse
relation and its two arguments are annotated for attribution.  The
attribute features are the Source or agent, the Type (assertion
propositions, belief propositions, facts and eventualities), scopal
polarity, and determinacy.  Scopal polarity is annotated on relations
and their arguments to identify cases when verbs of attribution are
negated on the surface but the negation takes scope over the embedded
clause.  An example is the sentence ``Having the dividend increases is
a supportive element in the market outlook {\it but I don't think it's
  a main consideration}.''  Here, the second
argument (the clause following {\it but}) is annotated with a
``Neg'' marker, meaning ``I think it's not a
main consideration.''

Wilson, Wiebe, and Hoffman~\shortcite{Wilson:2009} describe the
importance of correctly interpreting polarity in the context of
sentiment analysis, which is the task of identifying positive and
negative opinions, emotions and evaluations.  The authors have
established a set of features to distinguish between positive and
negative polarity and discuss the importance of correctly analyzing
the scope of the negation and the modality (e.g., whether the
proposition is asserted to be real or not real).

A major annotation effort for temporal and event expressions is the
TimeML specification language, which has been developed in the context
of reasoning for question answering \cite{SauriVP06}.  TimeML, which
includes modality annotation on events, is the basis for creating the
TimeBank and FactBank corpora \cite{Pustejovsky06,Sauri09}.  In
FactBank, event mentions are marked with their degree of factuality.

Recent work incorporating modality annotation includes work on
detecting certainty and uncertainty.  Rubin~\shortcite{Rubin07} describes a
scheme for five levels of certainty, referred to as Epistemic
modality, in news texts.  Annotators identify explicit certainty
markers and also take into account Perspective, Focus, and Time.
Focus separates certainty into facts and opinions, to include attitudes.  In our scheme,
focus would be covered by {\it want\/} and {\it belief\/} modality.
Also, separating focus and uncertainty can allow the annotation of
both on one trigger word.  Prabhakaran et al. \shortcite{Prabhakaran:2010} describe a scheme
for automatic committed belief tagging.  Committed belief indicates
the writer believes the proposition.  The authors use a previously
annotated corpus of committed belief, non-committed belief, and not
applicable~\cite{Diab:2009}, and derive features for machine
learning from parse trees.  The authors desire to combine their work
with FactBank annotation.

The CoNLL-2010 shared task~\cite{Farkas:2010} was about the detection
of cues for uncertainty and their scope.  The task was described as
``hedge detection,'' that is, finding statements which do not or
cannot be backed up with facts.  Auxiliary verbs such as {\it may\/},
{\it might\/}, {\it can\/}, etc. are one type of hedge cue.  The
training data for the shared task included the BioScope corpus~\cite{Szarvas:2008}, which
is manually annotated with negation and speculation cues and their
scope, and paragraphs from Wikipedia possibly containing hedge
information.  Our scheme also identifies cues in the form of triggers,
but our desired outcome is to cover the full range of modalities and not 
just certainty and uncertainty.  To identify scope, we use syntactic
parse trees, as was allowed in the CoNLL task.

The textual entailment literature includes modality annotation
schemes.  Identifying modalities is important to determine whether a
text entails a hypothesis.  Bar-Haim et al. \shortcite{Bar-Haim:2007} include polarity
based rules and negation and modality annotation rules.  The polarity
rules are based on an independent polarity lexicon \cite{Nairn:2006}.  The
annotation rules for negation and modality of predicates are based on
identifying modal verbs, as well as conditional sentences and modal
adverbials.  The authors read the modality off parse trees directly
using simple structural rules for modifiers.

Earlier work describing the difficulty of correctly translating
modality using machine translation includes~\cite{Sigurd:1994}
and~\cite{Murata:2005}.  Sigurd et al.~\shortcite{Sigurd:1994} write
about rule based frameworks and how using alternate grammatical
contructions such as the passive can improve the rendering of the
modal in the target language.  Murata et al.~\shortcite{Murata:2005}
analyze the translation of Japanese into English by several systems,
showing they often render the present incorrectly as the progressive.
The authors trained a support vector machine to specifically handle
modal contructions, while our modal annotation approach is a part of a
full translation system.

We now consider other literature, relating to tree-grafting and
machine translation.  Our tree-grafting approach builds on a
technique used for tree augmentation in Miller et 
al.~\shortcite{Miller:2000}, where
parse-tree nodes are augmented with semantic categories. 
In this earlier work, tree nodes were augmented with 
relations, while we augmented tree nodes with
modality and negation.  The parser is
subsequently retrained for both semantic and syntactic processing.
The semantic annotations were done manually by students 
who were provided a
set of guidelines and then merged with the syntactic trees
automatically.  In our work we tagged our corpus with entities,
modality, and negation automatically and then grafted them onto the
syntactic trees automatically, for the purpose of training a
statistical machine translation system.  An added benefit of the
extracted translation rules is that they are capable of producing
semantically-tagged Urdu parses, despite the fact that the training
data were processed by only an English parser and tagger.

Related work in syntax-based MT includes that of 
Huang and Knight~\shortcite{HuangKnight06}, where a
series of syntax rules are applied to a source language string to
produce a target language phrase structure tree.  The Penn English
Treebank~\cite{Marcus:93} is used as the source for the syntactic
labels and syntax trees are relabeled to improve translation quality.
In this work, node-internal and node-external information is used to
relabel nodes, similar to earlier work where structural context was
used to relabel nodes in the parsing domain~\cite{KleinManning03}.
Klein and Manning's methods include lexicalizing determiners and
percent markers, making more fine-grained VP categories, and marking
the properties of sister nodes on nodes. All of these labels are
derivable from the trees themselves and not from an auxiliary source.
Wang et al.~\shortcite{WangEtAl:2010} employ this type of node
splitting in machine translation and report a small increase in BLEU
score.

We use the methods described
in~\cite{ZollmannVenugopal:2006,Venugopal:2007} to induce synchronous
grammar rules, a process which requires phrase alignments and
syntactic parse trees.  Venugopal et al.~\shortcite{Venugopal:2007}
use generic non-terminal category symbols, as in~\cite{Chiang:2005},
as well as grammatical categories from the Stanford
parser~\cite{KleinManning03}.  Their method for rule induction generalizes
to any set of non-terminals.  We further refine this process
 by adding semantic notations onto the syntactic non-terminals
produced by a Penn Treebank trained parser, thus making the 
categories more informative.

In the parsing domain, the work of
Petrov and Klein~\shortcite{Petrov-Klein-2007:AAAI} is
related to the current work.  In this work,
rule splitting and rule merging are applied to refine parse trees during
machine learning.  Hierarchical splitting leads to the creation of
learned categories that have linguistic relevance, such as a breakdown
of a determiner category into two subcategories of determiners by
number, i.e., {\it{this}} and {\it{that}} group together as do
{\it{some}} and {\it{these}}.  We augment parse trees by category
insertion in cases where a semantic category is inserted as a node in
a parse tree, after the English side of the corpus has been parsed by
a statistical parser.

\section{SIMT Motivation}
\label{motivation}

As in many of the frameworks described above, the aim of the SIMT
effort was to provide a generalized framework for representing
structured semantic information, such as modality and negation.
Unlike many of the previous semantic annotation efforts (where the
emphasis tends to be on English), however, our approach is designed to
be directly integrated into a translation engine, with the goal of
translating highly divergent language pairs, such as Urdu and English.
As such, our choice of annotation scheme---illustrated in the
trigger-target example shown in Figure~\ref{modality-example}---was
based on a simplified structural representation that is general enough
to accommodate divergent modality/negation phenomena, easy for
language experts to follow, and straightforward to integrate into a
tree-grafting mechanism for MT.  Our objective is to investigate
whether incorporating this sort of information into machine
translation systems could produce better translations,
particularly in settings where only small parallel corpora are
available.

\begin{table}
\begin{center}
\begin{tabular}{lrrrrr} \hline
\multicolumn{2}{c}{} & \multicolumn{2}{c}{Urdu} & \multicolumn{2}{c}{English} \\
set & lines &  tokens &  types &  tokens & types\\ \hline 
training & 202k & 1.7M & 56k & 1.7M & 51k \\
dev & 981 & 21k & 4k & 19k & 4k \\
devtest & 883 & 22k & 4k & 19-20k & 4k \\
test & 1792 & 42k & 6k & 38-41k & 5k \\
\end{tabular}
\caption{The size of the various data sets used for the experiments in this paper including the training, development (dev), incremental test set (devtest) and blind test set (test).  The dev/devtest was a split of the NIST08 Urdu-English test set, and the blind test set was NIST09.}
\label{data-set-sizes}
\end{center}
\end{table}

\begin{figure}
\begin{center}
\includegraphics[width=\linewidth]{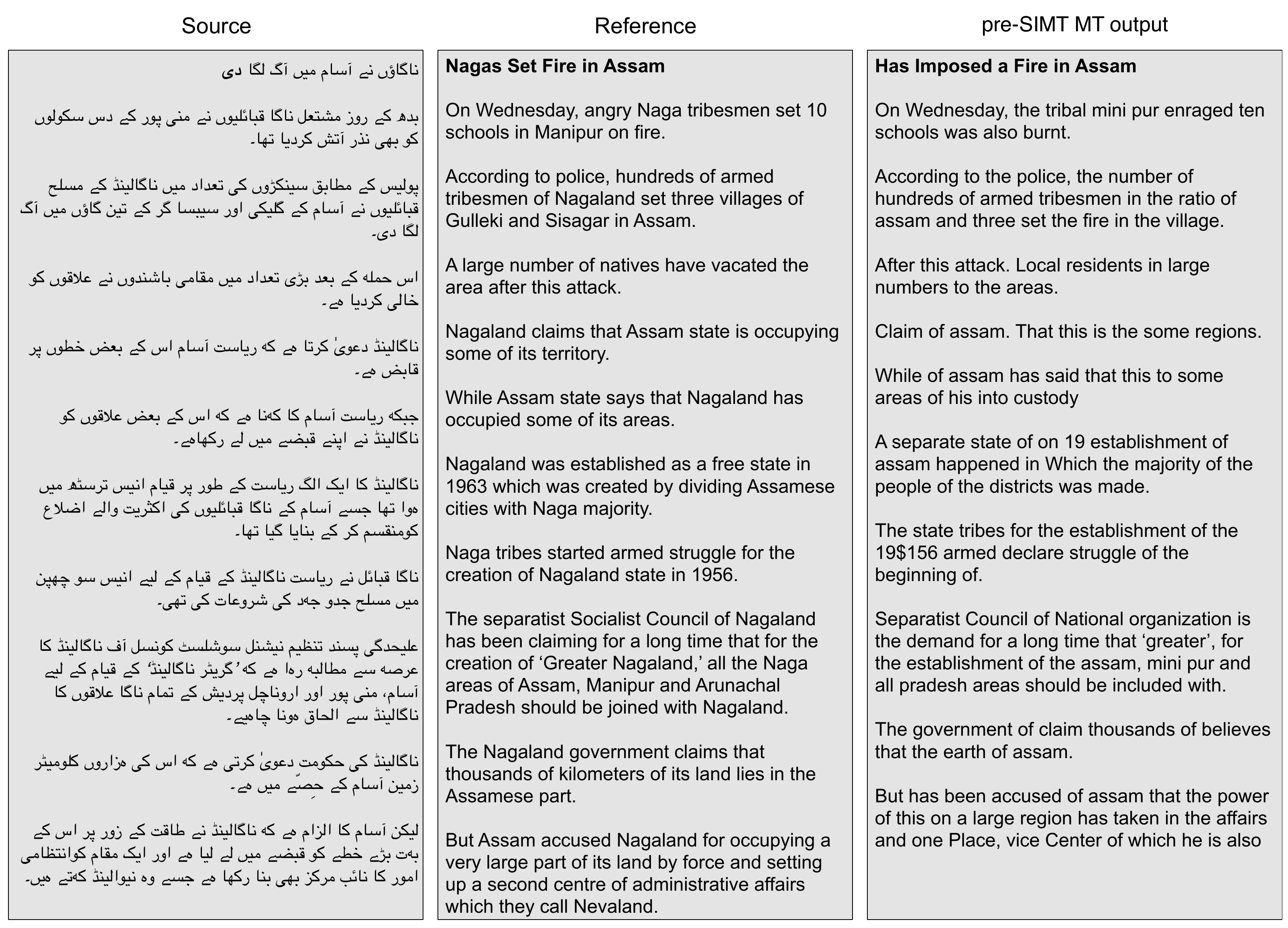}
\end{center}
\caption{An example of Urdu-English translation.  Shown are an Urdu
source document, a reference translation produced by a professional
human translator, and machine translation output from a phrase-based model (Moses) without linguistic information, which is representative of state-of-the-art MT quality before the SIMT effort.   }
\label{Urdu-example}
\end{figure}
It is informative to look at an example translation to understand the
challenges of translating important semantic elements when working
with a low-resource language pair. Figure~\ref{Urdu-example} shows an
example taken from the 2008 NIST Urdu-English translation task, and
illustrates the translation quality of a state-of-the-art Urdu-English
system (prior to the SIMT effort).  The small amount of training data
for this language pair (see Table~\ref{data-set-sizes}) results in
significantly degraded translation quality compared, e.g., to an
Arabic-English system that has more than 100 times the amount of
training data.

The output in Figure~\ref{Urdu-example} was produced using
Moses~\cite{Moses}, a state-of-the-art phrase-based MT system that by
default does not incorporate any linguistic information (e.g., syntax
or morphology or transliteration knowledge).  As a result, words that
were not directly observed in the bilingual training data were
untranslatable.  Names, in particular, are problematic.  For example,
the lack of translation for {\it Nagaland} and {\it Nagas} induces
multiple omissions throughout the translated text, thus producing
several instances where the {\it holder\/} of a claim (or {\it
  belief\/}) is missing.  This is because out of vocabulary words are
deleted from the Moses output.

\begin{figure}
\begin{center}
\includegraphics[height=3in]{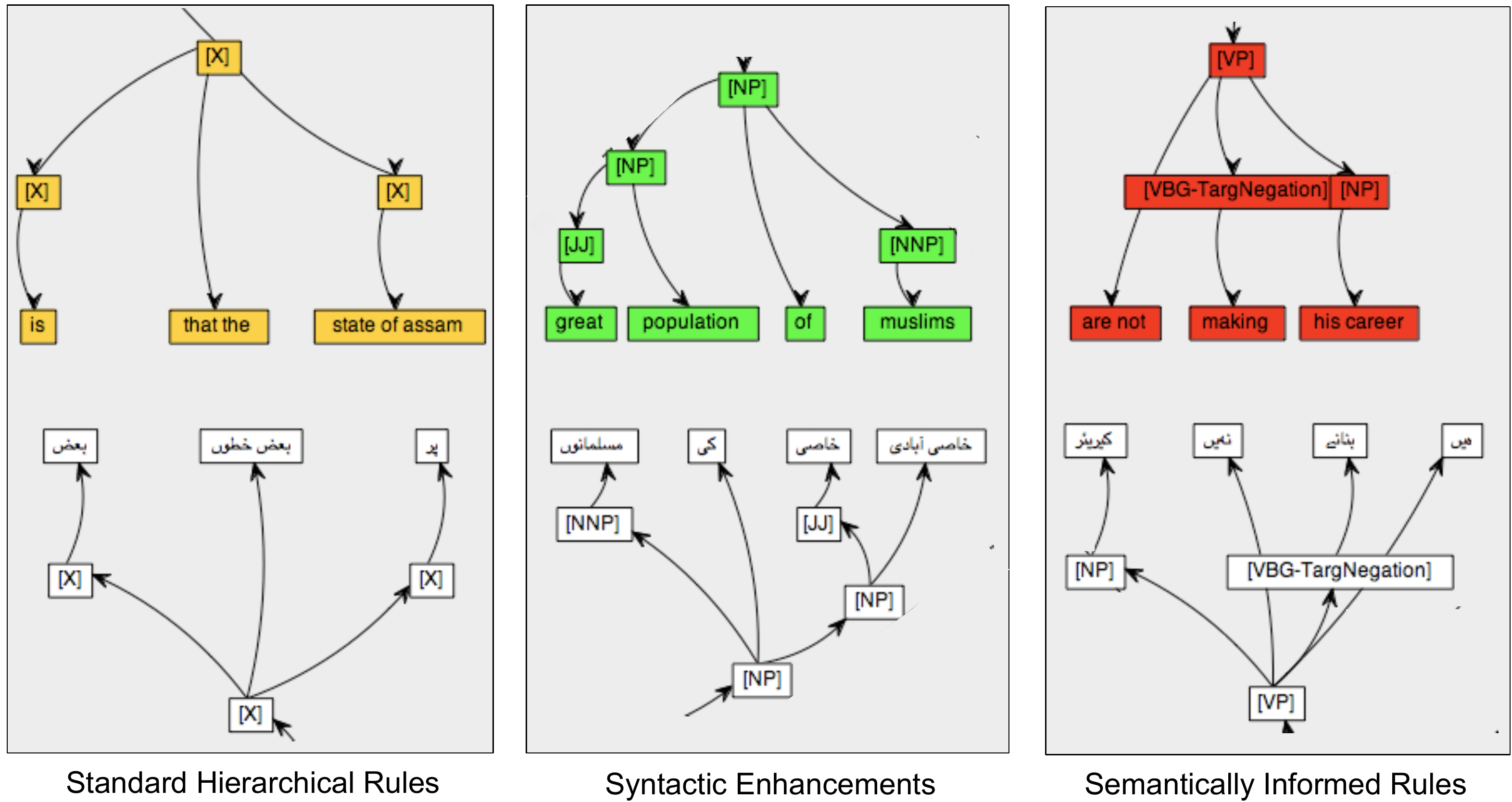}
\end{center}
\caption{The evolution of a semantically informed approach to our synchronous context free grammars (SCFGs).  At the start of summer the decoder used translation rules with a single generic non-terminal symbol, later syntactic categories were used, and by the end of the summer the translation rules included semantic elements such as modalities and negation, as well as named entities.}
\label{Evolution-to-SIMT-Joshua}
\end{figure}

We use syntactic and semantic tags as higher-order symbols inside
the translation rules used by the translation models.  Generic symbols
in translation rules (i.e., the non-terminal symbol ``X'') were
replaced with structured information at multiple levels of
abstraction, using a tree-grafting approach that we describe below.
Figure~\ref{Evolution-to-SIMT-Joshua} illustrates the evolution of the
translation rules that we used, first replacing ``X'' with grammatical
categories and then with categories corresponding to semantic units.

The semantic units that we examined in this effort were modalities and negation 
(indications that a statement represents something that has/hasn't taken place 
or is/isn't a belief or an intention) and  named entities
(such as people or organizations).  Other semantic units such as relations between
entities and events, were not part of this effort, but we believe they
could be similarly incorporated into the framework.  We chose to
examine semantic units that canonically exhibit two different
syntactic types: verbal, in the case of modality and negation, and nominal, in the 
case of named entities.

Although used in this effort, named entities were not the focus of our
research efforts in SIMT.  Rather, we focused on the development of an
annotation scheme for modality and negation and its use in MT,
while relying on a pre-existing HMM-based tagger derived from Identifinder
 ~\cite{Bikel:1999} to produce entity tags. Thus, the remainder of this paper will focus 
 on our modality/negation annotation scheme, two MN taggers produced by the effort, 
 and on the integration of semantics in the SIMT paradigm.

\section{Modality and Negation}
\label{modality-section}

Modality is an extra-propositional component of meaning.  In {\it John
  may go to NY\/}, the basic proposition is {\it John go to NY\/} and
the word {\it may\/} indicates modality and is called the {\it
  trigger\/} in our work. Van der Auwera and
Amman~\shortcite{VanDerAuweraAmman} define core cases of modality:
{\it John must go to NY\/} (epistemic necessity), {\it John might go
  to NY\/} (epistemic possibility), {\it John has to leave NY now\/}
(deontic necessity) and {\it John may leave NY now\/} (deontic
possibility).  Larreya~\shortcite{Larreya:2009} defines the core cases
slightly differently as {\it root\/} and {\it epistemic}.  Root
modality in Larreya's taxonomy includes physical modality ({\it He had
  to stop.  The road was blocked\/}) and deontic modality ({\it You
  have to stop\/}).  Epistemic modality includes problematic modality
({\it You must be tired\/}) and implicative modality ({\it You have to
  be mad to do that\/}).  Many
semanticists~\cite{Kratzer,VonFintelIatridou} define modality as
quantification over possible worlds. {\it John might leave NY\/} means
that there exist some possible worlds in which John leaves NY. Another
view of modality relates more to a speaker's attitude toward a
proposition~\cite{NirenburgMcShane,McShaneEtAl:2004}.
 
We incorporate negation as an inextricably intertwined component of
modality, using the term ``modality/negation (MN)'' to refer to our
resources (lexicons) and processes (taggers).  We adopt the view that
modality includes several types of attitudes that a speaker might have
(or not have) toward an event or state.  From the point of view of the reader or 
listener, modality might indicate factivity, evidentiality, or
sentiment.  Factivity is related to whether an event, state, or
proposition happened or didn't happen.  It distinguishes things that
happened from things that are desired, planned, or probable.
Evidentiality deals with the source of information and may provide
clues to the reliability of the information.  Did the speaker have
first hand knowledge of what he or she is reporting, or was it hearsay
or inferred from indirect evidence?  Sentiment deals with a speaker's
positive or negative feelings toward an event, state, or proposition.

Our project was limited to modal words and phrases---and their
negations---that are related to factivity.  However, beyond the core
cases of modality we include some aspects of speaker attitude such as
intent and desire.  We included these because they are often not
separable from the core cases of modality.  For example, {\it He had
  to go\/} may include the ideas that someone wanted him to go, that
he might not have wanted to go, that at some point after coercion he
intended to go, and that at some point he was able to
go~\cite{Larreya:2009}.

Our focus was on the eight modalities in Figure~\ref{8-modalities},
where P is a proposition (the {\it target\/} of the {\it triggering\/}
modality) and H is the holder (experiencer or cognizer of the
modality).  Some of the eight factivity-related modalities may overlap
with sentiment or evidentiality.  For example, {\it want\/} indicates
that the proposition it scopes over may not be a fact (it may just be
desired), but it also expresses positive sentiment toward the
proposition it scopes over.  We assume that sentiment and
evidentiality are covered under separate coding schemes, and that
words like {\it want\/} would have two tags, one for sentiment and one
for factivity.

\section{The Modality/Negation Annotation Scheme}
\label{modality-annotation-scheme}

The challenge of creating a MN annotation scheme was to deal with the
complex scoping of modalities with each other and with negation, while
at the same time creating a simplified operational procedure that
could be followed by language experts without special training. Below
we describe our MN annotation framework, including a set of linguistic
simplifications, and then we present our methodology for creation of a
publicly available MN lexicon.  The modality annotation scheme is
fully documented in a set of guidelines that were written with English
example sentences \cite{Baker:2010b}.  The guidelines can be used to
derive hand tagged evaluation data for English and they also include
a section that contains a set of Urdu trigger-word examples.

During the SCALE workshop, some Urdu speakers used the guidelines to
annotate a small corpus of Urdu by hand, which we reserved for future
work.  The Urdu corpus could be useful as an evaluation corpus for
automatically tagged Urdu, such as one derived from rule projection in
the Urdu-English machine translation system, a method we describe
further in section~\ref{joshua}. Also, although we did not annotate a
very large Urdu corpus, more data could be manually annotated to train
an automatic Urdu tagger in the future.

\subsection{Anatomy of Modality/Negation in Sentences} In sentences that
express modality, we identify three components: a trigger, a target,
and a holder.  The trigger is the word or string of words that
expresses modality or negation.  
The target is the event, state, or relation over which
the modality scopes.  The holder is the experiencer or cognizer
of the modality.  The trigger can be a word such as {\it should\/},
{\it try\/}, {\it able\/}, {\it likely\/}, or {\it want\/}.  It can
also be a negative element such as {\it not\/} or {\it n't\/}.  Often,
modality or negation is expressed without a lexical trigger.  For a typical
declarative sentence (e.g., {\it John went to NY\/}), the default
modality is strong belief when no lexical trigger is present.
Modality can also be expressed constructionally.  For example,
Requirement can be expressed in Urdu with a dative subject and
infinitive verb followed by a verb that means to happen or befall.   

\begin{figure}
\begin{itemize}
\item {\bf Requirement:} does H require P?
\item {\bf Permissive:} does H allow P?
\item {\bf Success:} does H succeed in P?
\item {\bf Effort:} does H try to do P?
\item {\bf Intention:} does H intend P?
\item {\bf Ability:} can H do P?
\item {\bf Want:} does H want P?
\item {\bf Belief}: with what strength does H believe P?
\end{itemize}
\caption{Eight Modalities Used for Tagging.  H stands for the Holder of the modality, and P is the proposition over which the modality has scope.}
\label{8-modalities}
\end{figure}

\subsection{Linguistic Simplifications for Efficient Operationalization}
\label{linguistic-simplifications}

Six linguistic simplifications were made for the sake of efficient
operationalization of the annotation task. The first linguistic
simplification deals with the scope of modality and negation.  The
first sentence below indicates scope of modality over negation.  The
second indicates scope of negation over modality:
\begin{itemize}
\item He tried not to criticize the president.
\item He didn't try to criticize the president.
\end{itemize}

The interaction of modality with negation is complex, but was
operationalized easily in the menu of thirteen choices shown in
Figure~\ref{13-menu-choices-for-modality}.  First consider the case
where negation scopes over modality.  Four of the thirteen choices are
composites of negation scoping over modality.  For example, the
annotators can choose {\it try\/} or {\it not try\/} as two separate
modalities.  Five modalities (Require, Permit, Want, Firmly Believe,
and Believe) do not have a negated form.  For three of these
modalities (Want, Firmly Believe, and Believe), this is because they
are often transparent to negation.  For example, {\it I do not believe
  that he left NY\/} sometimes means the same as {\it I believe he
  didn't leave NY\/}.  Merging the two is obviously a simplification,
but it saves the annotators from having to make a difficult decision.

The second linguistic simplification is related to a duality in
meaning between {\it require\/} and {\it permit\/}.  Not requiring P
to be true is similar in meaning to permitting P to be false.  Thus,
annotators were instructed to label {\it not require P to be true\/}
as {\it Permit P to be false\/}.  Conversely, {\it not Permit P to be
  true\/} was labeled as {\it Require P to be false\/}. 

After the annotator chooses the modality, the scoping of modality over
negation takes place as a second decision.  For example, for the
sentence {\it John tried not to go to NY\/}, the annotator first
identifies {\it go\/} as the target of a modality and then
chooses {\it try\/} as the modality.  Finally, the
annotator chooses {\it false\/} as the polarity of the target.

\begin{figure}
\begin{itemize}
\item H requires [P to be true/false]
\item H permits [P to be true/false]
\item H succeeds in [making P true/false]
\item H does not succeed in [making P true/false]
\item H is trying [to make P true/false]
\item H is not trying [to make P true/false]
\item H intends [to make P true/false]
\item H does not intend [to make P true/false]
\item H is able [to make P true/false]
\item H is not able [to make P true/false]
\item H wants [P to be true/false]
\item H firmly believes [P is true/false]
\item H believes [P may be true/false] 
\end{itemize}
\caption{Thirteen Menu Choices for Modality/Negation Annotation.}
\label{13-menu-choices-for-modality}
\end{figure}

The third simplification relates to entailments between modalities.
Many words have complex meanings that include components of more than
one modality.  For example, if one managed to do something, one tried
to do it and one probably wanted to do it.  Thus, annotators were
provided a specificity-ordered modality list in
Figure~\ref{13-menu-choices-for-modality}, and were asked to choose the
first applicable modality.  We note that this list corresponds to two
independent ``entailment groupings,'' ordered by specificity:
\begin{itemize}
\item \{{\it requires\/} $\rightarrow$ {\it permits\/}\}
\item \{{\it
  succeeds\/} $\rightarrow$ {\it tries\/} $\rightarrow$ {\it
  intends\/} $\rightarrow$ {\it is able\/} $\rightarrow$ {\it
  wants\/}\}
\end{itemize}
Inside the entailment groupings, the ordering
corresponds to an entailment relation, e.g., {\it succeeds\/} can only
occur if {\it tries\/} has occurred.  Also, the \{{\it requires
  $\rightarrow$ $\ldots$ \/}\} entailment grouping is taken to be more
specific than (ordered before) the \{{\it succeeds $\rightarrow$
  $\ldots$ \/}\} entailment grouping.  Moreover, both entailment
groupings are taken to be more specific than {\it believes\/},
which is not in an entailment relation with any of the other
modalities.

The fourth simplification, already mentioned above, is that sentences
without an overt trigger word are tagged as {\it firmly believes}.  This
heuristic works reasonably well for the types of documents we were
working with, although one could imagine genres such as fiction in
which many sentences take place in an alternate possible world
(imagined, conditional, or counterfactual) without explicit marking.

The fifth linguistic simplification is that we did not require
annotators to mark nested modalities.  For a sentence like {\it He
might be able to go to NY\/} the target word {\it go\/} is marked as
ability, but {\it might\/} is not annotated for Belief modality. 
This decision was based on time limits on the annotation task;
there was not enough time for annotators to deal with syntactic
scoping of modalities over other modalities.

Finally, we did not mark the holder H because of the short time frame for
workshop preparation.  We felt that identifying the triggers and
targets would be most beneficial in the context of machine
translation.

\subsection{The English Modality/Negation Lexicon}
\label{English-Modality-Lexicon}

Using the framework described above, we created a MN lexicon that was
incorporated into a MN tagging scheme to be described below in
Section~\ref{Automatic-Modality-Annotation}. Entries in the MN lexicon
consist of: (1) A string of one or more words: for example, {\it
  should\/} or {\it have need of\/}.  (2) A part of speech for each
word: the part of speech helps us avoid irrelevant homophones such as
the noun {\it can\/}.  (3) A MN designator: one of the thirteen
modality/negation cases described above.  (4) A head word (or {\it
  trigger\/}): the primary phrasal constituent to cover cases where an
entry is a multi-word unit, e.g., the word {\it hope\/} in {\it hope
  for\/}. (5) One or more subcategorization codes.

We produced the full English MN lexicon semi-automatically. First, we gathered a small seed list of MN trigger words and phrases from our modality annotation manual \cite{Baker:2010b}. Then, we expanded this small list of MN trigger words by running an online search for each of the words, specifically targeting free online thesauri (e.g., thesaurus.com), to find both synonymous and antonymous words.  From these we manually selected the words we thought triggered modality (or their corresponding negative variants) and filtered out words that we thought didn't trigger modality. The resulting list of MN trigger words and phrases contained about 150 lemmas.

We note that most intransitive LDOCE codes were not applicable to
modality/negation constructions.  For example, {\it hunger\/} (in the {\it
Want\/} modality class) has a modal reading of ``desire'' when
combined with the preposition {\it for\/} (as in {\it she hungered for
a promotion\/}), but we do not consider it to be modal when it is used in the
somewhat archaic sentence {\it He hungered\/}, meaning that he did not
have enough to eat.   Thus the LDOCE code \verb|I| associated
with the verb {\it hunger\/} was hand-changed to \verb|I-FOR|.
There were 43 such cases.  Once the LDOCE codes
were hand-verified (and modified accordingly), the mapping to
subcategorization codes was applied.

We note that most intransitive LDOCE codes were not applicable to
modality/negation constructions.  For example, {\it hunger\/} (in the {\it
Want\/} modality class) has a modal reading of ``desire'' when
combined with the preposition {\it for\/} (as in {\it she hungered for
a promotion\/}), but not in its pure intransitive form (e.g., {\it he
hungered all night\/}).  Thus the LDOCE code \verb|I| associated
with the verb {\it hunger\/} was hand-changed to \verb|I-FOR|.
There were 43 such cases.  Once the LDOCE codes
were hand-verified (and modified accordingly), the mapping to 
subcategorization codes was applied.  

The MN lexicon is publicly available at
\url{http://www.umiacs.umd.edu/~bonnie/ModalityLexicon.txt}.  An
example of an entry is given in Figure~\ref{need-entry},
for the verb {\it need\/}.
\begin{figure}
\begin{center}
\begin{tabular}{lp{4.3in}}
{\bf String:}&Need\\
{\bf Pos:}&VB\\
{\bf Modality:}&Require \\
{\bf Trigger:}&Need\\
{\bf Subcat:}& {\bf V3-passive-basic} --
More citizens are needed to vote.\\
{\bf Subcat:}& {\bf V3-I3-basic} --
The government will need to work continuously
for at least a year.  We will need them to work continuously.\\
{\bf Subcat:}& {\bf T1-monotransitive-for-V3-verbs} --
We need a Sir Sayyed again to maintain this sentiment.\\
{\bf Subcat:}& {\bf T1-passive-for-V3-verb} --
Tents are needed.\\
{\bf Subcat:}& {\bf Modal-auxiliary-basic} --
He need not go.
\end{tabular}
\end{center}
\caption{Modality Lexicon Entry for {\it need\/}.}

\label{need-entry}
\end{figure}

\section{Automatic Modality/Negation Annotation}
\label{Automatic-Modality-Annotation}

A MN tagger produces text or structured text in which modality or negation
triggers and/or targets are identified.  Automatic identification of
the holders of modalities was beyond the scope of our project because
the holder is often not explicitly stated in the sentence in which the
trigger and target occur.  This section describes two types of
MN taggers---one that is string-based and one that is
structure-based.

\subsection{The String-based English Modality/Negation Tagger}
\label{string-based-english-modality-tagger}

The string-based tagger operates on text that has been tagged with
parts of speech by 
a Collins-style statistical parser~\cite{Miller:1998}.
The tagger marks spans of
words/phrases that exactly match MN trigger words in the
MN lexicon described above, and that exactly match the same
parts of speech. This tagger identifies the target of each modality/negation
using the heuristic of tagging the next non-auxiliary verb to the
right of the trigger.  Spans of words can be tagged multiple times
with different types of triggers and targets.

We found the string-based modality/negation tagger to produce output
that matched about 80\% of the sentence-level tags produced by our
structure-based tagger, the results of which are described next. While
string-based tagging is fast and reasonably accurate in practice, we
opted to focus on the in-depth analysis of modality/negation of our
SIMT results using the more accurate structure-based tagger.

\subsection{The Structure-based English Modality/Negation Tagger}
\label{structured-based-english-modality-tagger}

The structure-based MN tagger operates on text that has been
parsed~\cite{Miller:1998}.  We used a version of the parser that
produces flattened trees.  In particular, the flattener deletes VP
nodes that are immediately dominated by VP or S and NP nodes that are
immediately dominated by PP or NP.  The parsed sentences are processed
by TSurgeon rules.  Each TSurgeon rule consists of a pattern and an
action.  The pattern matches part of a parse tree and the action
alters the parse tree.  More specifically, the pattern finds a
MN trigger word and its target and the action inserts tags such
as {\tt TrigRequire} and {\tt TargRequire} for triggers and targets
for the modality Require.  Figure~\ref{structure-based-tagger-output}
shows output from the structure-based MN tagger. (Note that the
sentence is disfluent: {\it Pakistan which could not reach semi-final,
  in a match against South African team for the fifth position
  Pakistan defeated South Africa by 41 runs.\/}) The example shows
that {\it could\/} is a trigger for the Ability modality and {\it
  not\/} is a trigger for negation.  {\it Reach\/} is a target for
both Ability and Negation, which means that it is in the category of
``H is not able [to make P true/false]'' in our coding scheme.  {\it
  Reach\/} is also a trigger for the Succeed modality and {\it
  semi-final\/} is its target.

\begin{figure}
\begin{small}
\begin{verbatim}
(TOP
 (S
  (NP 
   (NNP Pakistan) 
   (SBAR (WDT which) 
    (S (MD TrigAble could) 
       (RB TrigNegation not) 
       (VB B TargAble TrigSucceed 
        TargNegation reach) 
       (ADJP 
        (JJ TargSucceed semi-final)) 
        (, ,) 
       (PP (IN in) (DT a) 
           (NN match) (PP (IN against) 
           (ADJP (JJ South) (JJ African))
            (NN team)) 
           (PP (IN for) (DT the)
               (JJ fifth) (NN position))
           (NP (NNP Pakistan)))))) 
  (VB D defeated) 
  (NP (NNP South) (NNP Africa)) 
  (PP (IN by) (CD 41) (NNS runs)) (. .)))
\end{verbatim}
\end{small}
\caption{Sample output from the structure-based MN tagger.}
\label{structure-based-tagger-output}
\end{figure}

The TSurgeon patterns are automatically generated from the verb class
codes in the MN lexicon along with a set of fifteen templates.  Each
template covers one situation such as the following: the target is the
subject of the trigger; the target is the direct object of the
trigger; the target heads an infinitival complement of the trigger;
the target is a noun modified by an adjectival trigger, etc.  The verb
class codes indicate which templates are applicable for each trigger
word.  For example, a trigger verb in the transitive class may use two
target templates, one in which the trigger is in active voice and the
target is a direct object ({\it need tents\/}) and one in which the
trigger is in passive voice and the target is a subject ({\it tents
  are needed\/}).

In developing the TSurgeon rules, we first conducted a corpus analysis
for forty of the most common trigger words in order to identify and
debug the most broadly applicable templates.  We then used LDOCE to
assign verb classes to the remaining verbal triggers in the MN
lexicon, and we associated one or more debugged templates with each
verb class.  In this way, the initial corpus work on a limited number
of trigger words was generalized to a longer list of trigger words.
Because the TSurgeon patterns are tailored to the flattened structures
produced by our parser, it is not easily ported to new parser
outputs. However, the MN lexicon itself is portable.  Switching parsers
would entail writing new TSurgeon templates, but the trigger words in
the MN lexicon would still be automatically assigned to templates
based on their verb classes.

The following example shows an example of a TSurgeon pattern-action
pair for a sentence like {\it They were required to provide tents\/}.
The pattern-action pair is intended to be used after a pre-processing
stage in which labels such as ``VoicePassive'' and ``AUX'' have been
assigned.  ``VoicePassive'' is inserted by a pre-processing TSurgeon
pattern because, in some cases, the target of a passive modality
trigger word is in a different location from the target of the
corresponding active modality trigger word.  ``AUX'' is inserted during
pre-processing to distinguish auxiliary uses of {\it have\/} and {\it
  be\/} from their uses as main verbs.  The pattern portion of the
pattern-action pair matches a node with label VB that is not already
tagged as a trigger and that is passive and dominates the string
"required".  The VB node is also a sister to an S node, and the S node
dominates a VB that is not an auxiliary ({\it provide\/} in this
case).  The action portion of the pattern-action pair inserts the
string ``TargReq'' as the second daughter of the second VB and inserts
the string ``TrigReq'' as the second daughter of the first VB.

\begin{small}
\begin{verbatim}
 VB=trigger !< /^Trig/ < VoicePassive < required  $.. 
    (S < (VB=target !< AUX))
 insert (TargReq) >2 target
 insert (TrigReq) >2 trigger
\end{verbatim}
\end{small}

Verb-specific patterns such as this one were generalized in order to
gain coverage of the whole modality lexicon.  The specific lexical
item, {\it required\/}, was replaced with a variable, as were the
labels ``TrigReq'' and ``TargReq.''  The pattern was then given a
name, V3-passive-basic, where V3 is a verb class tag from LDOCE
(described above in Section~\ref{English-Modality-Lexicon}) for verbs
that take infinitive complements.  We then looked up the LDOCE verb
class labels for all of the verbs in the modality lexicon.  Using this
information, we could then generate a set of new, verb-specific
patterns for each V3 verb in the modality lexicon.

\subsection{Evaluating the Effectiveness of Structure-based MN Tagging}
\label{modality-tagging-evaluation}

We performed a manual inspection of the structure-based tagging
output.  We calculated precision by examining 229 instances of
modality triggers that were tagged by our tagger from the English side
of the NIST 09 MTEval training sentences.  We analyzed precision in
two steps, first checking for the correct syntactic position of the
target and then checking the semantic correctness of the trigger and
target.  For 192 of the 229 triggers (around 84\%), the targets were
tagged in the correct syntactic location.  

For example, for the sentence {\it A solution must be found to this
  problem\/} shown in Figure~\ref{embedded-targ-example}, the word {\it
  must\/} is a modality trigger word, and the correct target is the
first non-auxiliary verb heading a verb phrase that is contained in
the syntactic complement of {\it must\/}.  The syntactic complement of
{\it must\/} is the verb phrase {\it be found to this problem\/}.  The
syntactic head of that verb phrase, {\it be\/}, is skipped because it
is an auxiliary verb.  The correct (embedded) target {\it found\/} 
is the head of the syntactic complement of {\it be\/}.

\begin{figure}
\begin{small}
\begin{verbatim}
(S (NP (DT A) (NN solution))
   (VP (MD-TrigBelief must) 
       (VP (VB be) 
           (VP (VBN-TargBelief found) 
               (PP (TO to) (NP (DT this) (NN problem)))))))
\end{verbatim}
\end{small}
\caption{Example of Embedded Target Head {\it found\/} inside VP {\it must be found\/}.}
\label{embedded-targ-example}
\end{figure}

The 192 modality instances with structurally correct targets do not
all have semantically correct tags.  In the example above, {\it
must\/} is tagged as {\tt TrigBelief}, where the correct tag would
be {\tt TrigRequire\/}.  Also, because the MN lexicon was used without
respect to word sense, words were sometimes erroneously identified as
triggers.  This includes non-modal uses of {\it work\/} (work with
refugees), {\it reach\/} (reach a destination), and {\it attack\/}
(attack a physical object), in constrast to modal uses of these words:
{\it work for peace\/} (effort), {\it reach a goal\/} (succeed), and
{\it attack a problem\/} (effort).  Fully correct tagging of modality
would need to include word sense disambiguation.

For 37 of the 229 triggers we examined, a target was not tagged in the
correct syntactic position.  In 12 of 37 incorrectly tagged instances
the targets are inside compound nouns or coordinate structures (NP or
VP), which are not yet handled by the modality tagger.  The remaining
25 of the 37 incorrectly tagged instances had targets that were lost
because the tagger does not yet handle all cases of nested modalities.  
Nested modalities occur in sentences like {\it They did not want to succeed
in winning\/} where the target words {\it want\/} and {\it
succeed\/} are also modality trigger words.  Proper treatment of
nested modalities requires consideration of scope and compositional
semantics.  

Nesting was treated in two steps.  First, the modality tagger marked
each word as a trigger and/or target.  In {\it They did not want to
  succeed in winning\/}, {\it not\/} is marked as a trigger for
negation, {\it want\/} is marked as a target of negation and a trigger
of wanting, {\it succeed\/} is marked as a trigger of succeeding and a
target of wanting, and {\it win\/} is marked as a target of
succeeding.  The second step in the treatment of nested modalities
occurs during tree grafting, where the meanings of the nested
modalities are composed.  The tree grafting program correctly composes
some cases of nested modalities.  For example, the tag {\tt TrigAble}
composed with {\tt TrigNegation} results in the target tag {\tt
TargNOTAble}, as shown in Figure~\ref{tree-grafting-composition}.
In other cases, where compositional semantics are not yet
accommodated, the tree grafting program removed target labels from the
trees, and those cases were counted as incorrect for the purpose of
this evaluation.

\begin{figure}
\begin{small}
\begin{verbatim}
(S 
  (NP (EX there)) 
  (VP (VBZ is) 
      (NP (NP (DT no) (NN difficulty))
          (SBAR (WHNP (WDT which))
            (S (VP (MD-TrigAble can) 
                   (RB-TrigNegation not) 
                   (VP (VB be) 
                       (VP-TargNOTAble (VBN-TargNOTAble solved)))))))))
\end{verbatim}
\end{small}
\caption{Example of Modality Composed with Negation: TrigAble and TrigNegation combine to form NOTAble.}
\label{tree-grafting-composition}
\end{figure}

In the 229 instances that we examined, there were 14 in which a light
verb or noun was the correct syntactic target, but not the correct
semantic target.  {\it Decision\/} would be a better target than {\it
taken\/} in {\it The decision {\bf should} be {\bf taken} on delayed
cases on the basis of merit.\/} We counted sentences with
semantically light targets as correct in our evaluation because our
goal was to identify the synactic head of the target.  The semantics
of the target is a general issue, and we often find lexico-syntactic
fluff between the trigger and the most semantically salient target in
sentences like {\it We succeeded in our goal of winning the war\/}
where ``success in war'' is the salient meaning.

With respect to recall, the tagger primarily missed special forms of
negation in noun phrases and prepositional phrases: {\it There was
{\bf no} place to seek shelter.\/}; {\it The buildings should be
reconstructed, {\bf not} with RCC, but with the wood and steel
sheets.\/} More complex constructional and phrasal triggers were
also missed: {\it President Pervaiz Musharraf has said that he will
{\bf not rest unless} the process of rehabilitation is completed.\/}
Finally, we discovered some omissions from our MN lexicon: {\it
It is not {\bf possible} in the middle of winter to re-open the
roads.\/}  Further annotation experiments are planned, which will
be analyzed to close such gaps and update the lexicon as appropriate.

Providing a quantitative measure of recall was beyond the scope of
this project.  At best we could count instances of sentences
containing trigger words that were not tagged.  We are also aware of
many cases of modality that were not covered such as the modal uses of
the future tense auxiliary {\it will\/} as in {\it That'll be John\/}
(conjecture), {\it I'll do the dishes\/} (volition), {\it He won't do
it\/} (non-volition), and {\it It will accommodate five\/}
(ability)~\cite{Larreya:2009}.  However, because of the complexity and
subtlety of modality and negation, it would be impractical to count
every clause (such as the {\it not rest unless\/} clause above) that
had a nuance of non-factivity.

\section{Semantically-Informed Syntactic MT}
\label{joshua}

This section describes the incorporation of our structured-based
MN tagging into an Urdu-English machine-translation system using
{\it tree grafting\/} for combining syntactic symbols with semantic
categories (e.g. modality/negation).  We note that a {\it de
  facto\/} Urdu MN tagger resulted from identifying the English
MN trigger and target words in a parallel English-Urdu corpus,
and then projecting the trigger and target labels to the corresponding
words in Urdu syntax trees.

\subsection{Refinement of Translation Grammars with Semantic Categories}

We used synchronous context free grammars (SCFGs) as the underlying
formalism for our statistical models of translation.  SCFGs provide a
convenient and theoretically grounded way of incorporating linguistic
information into statistical
models of translation, by specifying grammar rules with syntactic
non-terminals in the source and target languages.  We refine the set
of non-terminal symbols so that they not only include syntactic
categories, but also semantic categories.

Chiang \shortcite{Chiang:2005} re-popularized the use of SCFGs for
machine translation, with the introduction of his hierarchical
phrase-based machine translation system, Hiero.  Hiero uses grammars
with a single non-terminal symbol ``X'' rather than using
linguistically informed non-terminal symbols.  When moving to
linguistic grammars, we use Syntax Augmented Machine Translation
(SAMT) developed by Venugopal et al.  \shortcite{Venugopal:2007}.  In
SAMT the ``X'' symbols in translation grammars are replaced with
nonterminal categories derived from parse trees that label the English
side of the Urdu-English parallel corpus.\footnote{For non-constituent
  phrases, composite CCG-style categories are
  used~\cite{Steedman:1999}.}  We refine the syntactic categories by
combining them with semantic categories.  Recall this progression was
illustrated in Figure~\ref{Evolution-to-SIMT-Joshua} on page~\pageref{Evolution-to-SIMT-Joshua}.

We extracted SCFG grammar rules containing modality,
negation and named entities using an extraction procedure that requires parse trees
for one side of the parallel corpus. While it is assumed that these
trees are labeled and bracketed in a syntactically motivated fashion,
the framework places no specific requirement on the label inventory.
We take advantage of this characteristic by providing the rule
extraction algorithm with augmented parse trees containing syntactic
labels that have semantic annotations grafted
onto them so that they additionally express semantic information.

\begin{figure}
\begin{center}
\includegraphics[width=\linewidth]{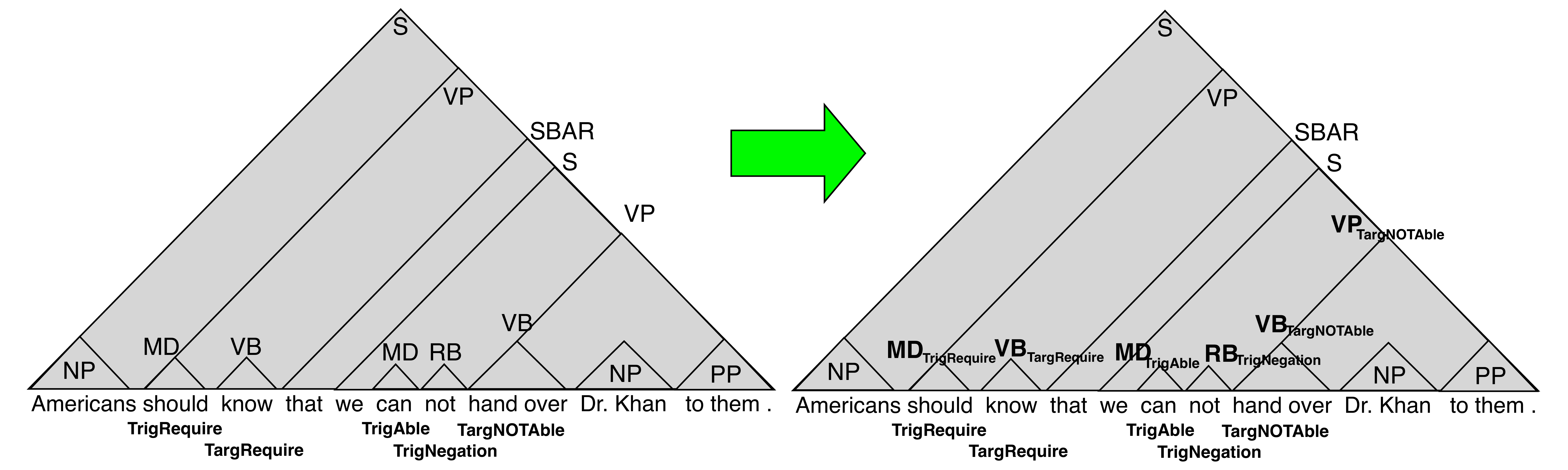}
\end{center}
\caption{A sentence on the English side of the bilingual parallel training corpus is parsed with a syntactic parser, and also tagged with our modality tagger.  The tags are then {\it grafted\/} onto the syntactic parse tree to form new categories like VP-TargNOTAble and VP-TargRequire.  Grafting happens prior to extracting translation rules, which happens normally except for the use of the augmented trees.}
\label{Parse-Tree-for-Sample-Sentence}
\end{figure}

Our strategy for producing semantically-grafted parse trees involves
three steps:
\begin{enumerate}[topsep=0pt, partopsep=0pt, itemsep=0.5pt, parsep=0.5pt]
\item The English sentences in the parallel training data are parsed
  with a syntactic parser.  In our work, we used the lexicalized
  probabilistic context free grammar parser provided by Basis
  Technology Corporation.
\item The English sentences are MN-tagged by the
  system described above and named-entity-tagged by the Phoenix
  tagger~\cite{richman-schone:2008:ACLMain}.
\item The modality/negation and entity markers are grafted onto the syntactic parse trees using a tree-grafting procedure.  The grafting procedure was implemented
as a part of the SIMT effort.  Details are further spelled out in Section~\ref{algorithm}.
\end{enumerate}

Figure~\ref{Parse-Tree-for-Sample-Sentence} illustrates how
modality tags are grafted onto a parse tree.  Note that while we focus the discussion here on the
modality and negation, our framework is general and we were able to incorporate other semantic elements (specifically, named entities) into the SIMT effort.

Once the semantically-grafted trees have been produced for the
parallel corpus, the trees are presented, along with word alignments
(produced by the  Berkeley aligner), to the rule extraction
software to extract synchronous grammar rules that are both
syntactically and semantically informed.  These grammar rules are used
by the decoder to produce translations.  In our experiments, we used
the Joshua decoder \cite{Li:2009}, the SAMT grammar extraction
software \cite{Venugopal2009}, and special purpose-built tree-grafting
software.

\begin{figure}
\begin{center}
\includegraphics[width=3.5in]{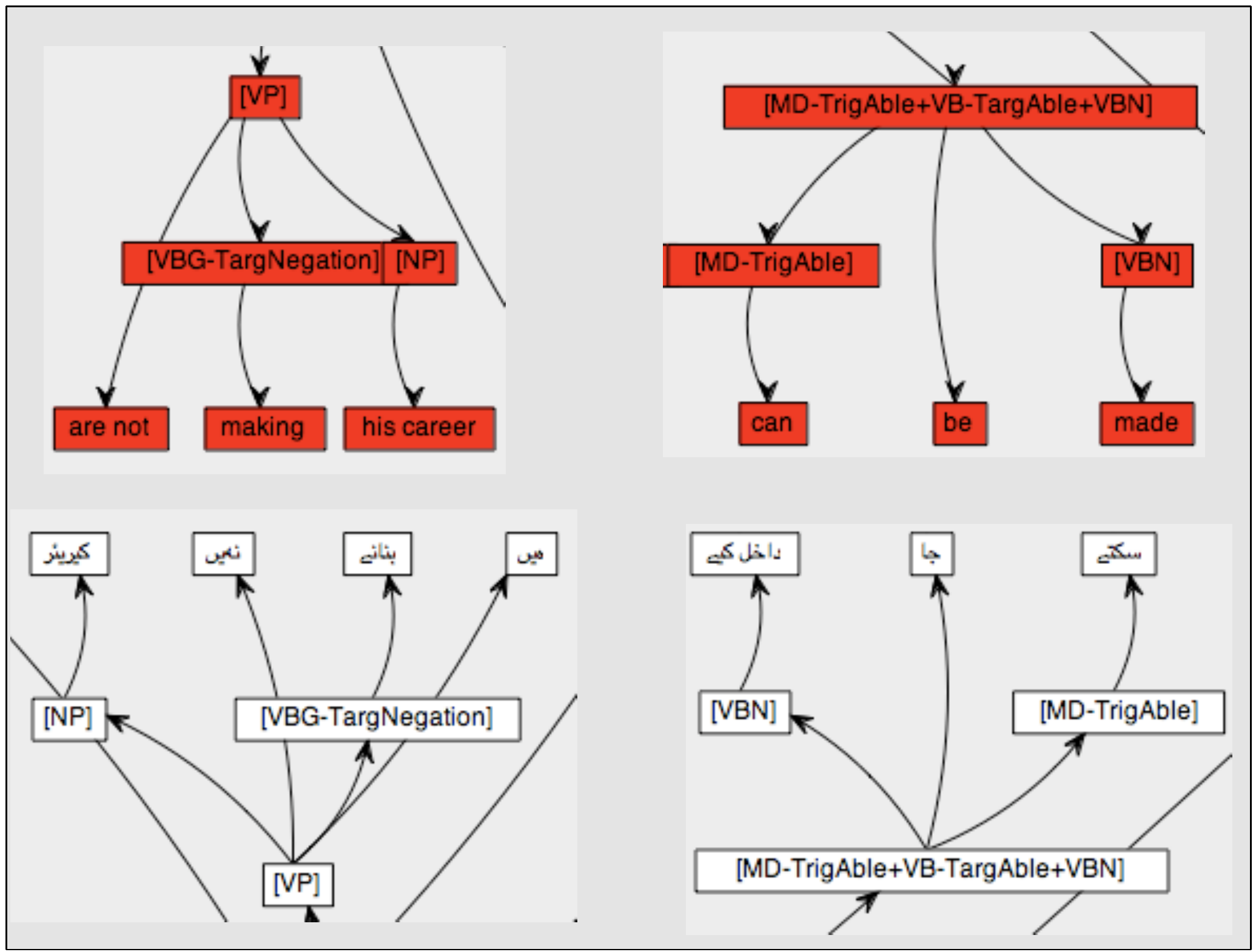}
\caption{Example translation rules with tags for modality, negation, and entities combined with syntactic categories.}\label{derivation-with-modalities}
\end{center}
\end{figure}

Figure~\ref{derivation-with-modalities} shows example semantic rules
that are used by the decoder.  
The verb phrase rules are augmented with
modality and negation,
taken from the semantic categories listed in
Table~\ref{Modalitytable}.  Because these get
marked on the Urdu source as well as the English translation,
semantically enriched grammars also act as very simple named entity or
MN taggers for Urdu.  However, only entities, modality,
and negation that occurred in the parallel training corpus are
marked in the output.

\subsection{Tree-Grafting Algorithm}
\label{algorithm}

The overall scheme of our tree-grafting algorithm is to match semantic
tags to syntactic categories.  There are two inputs to the process.
Each is derived from a common text file of sentences.  The first input
is a list of standoff annotations for the semantically tagged word
sequences in the input sentences, indexed by sentence number.  The
second is a list of parse trees for the sentences in Penn Treebank
format, indexed by sentence number.
 
Table~\ref{Modalitytable} lists the modality/negation types that were
produced by the MN tagger.  For example, the sentence {\it The
  students are able to swim\/} is tagged as {\it The students are
  $\langle$TrigAble able$\rangle$ to $\langle$TargAble
  swim$\rangle$\/}.  The distinction between ``Negation'' and ``NOT''
corresponds to the difference between negation that is inherently
expressed in the triggering lexical item and negation that is
expressed explicitly as a separate lexical item.  Thus, {\it I
  achieved my goal\/} is tagged ``Succeed'' and {\it I did not achieve
  my goal\/} is tagged as ``NOTSucceed,'' but {\it I failed to win\/}
is tagged as ``SucceedNegation,'' and {\it I did not fail to win\/} is
tagged as ``NOTSucceedNegation.''

\begin{table}
\begin{center}
\begin{small}
\begin{tabular}{ll}\\
Require& NOTRequire\\
Permit& NOTPermit\\
Succeed & NOTSucceed\\
SucceedNegation & NOTSucceedNegation\\
Effort& NOTEffort\\
EffortNegation & NOTEffortNegation\\
Intend& NOTIntend\\
IntendNegation& NOTIntendNegation\\
Able& NOTAble\\
AbleNegation& NOTAbleNegation\\
Want& NOTWant\\
WantNegation& NOTWantNegation\\
Belief& NOTBelief\\
BeliefNegation& NOTBeliefNegation\\
Firm\_Belief& NOTFirm\_Belief\\
Firm\_BeliefNegation& NOTFirm\_BeliefNegation\\
Negation & \\
\end{tabular}
\end{small}
\end{center}
\caption{Modality Tags with their Negated Versions. Note that {\it Require\/} and {\it Permit\/} are in a dual relation, and thus RequireNegation is represented as NOTPermit and PermitNegation is represented as NOTRequire.}
\label{Modalitytable}
\end{table}

The tree-grafting algorithm proceeds as follows. For each tagged
sentence, we iterate over the list of semantic tags.  For each
semantic tag, there is an associated word or sequence of words.  For
example, the modality tag TargAble may tag the word {\it swim}.

For each semantically tagged word, we find the parent node in the
 corresponding syntactic parse tree that dominates that word.  For a
 word sequence, we find and compare the parent nodes for all of the
 words.  Each node in the syntax tree has a category label.  The
 following tests are then made and tree grafts applied:
\begin{itemize}
\item If there is a single node in the parse tree that dominates all
  and only the words with the semantic tag, graft the name of the
  semantic tag onto the highest corresponding syntactic constituent in
  the tree.  For example, in
  Figure~\ref{Parse-Tree-for-Sample-Sentence}, which shows the
  grafting process for modality tagging, the semantic tag TargNOTAble
  that ``hand over'' receives is grafted onto the VB node that
  dominates all and only the words ``hand over.'' Then the semantic
  tag TargNOTAble is passed up the tree to the VP node, which is the
  highest corresponding syntactic constituent.
\item If the semantic tag corresponds to words that are adjacent
  daughters in a syntactic constituent, but less than the full
  constituent, insert a node dominating those words into the parse
  tree, as a daughter of the original syntactic constituent.  The name
  of the semantic tag is grafted onto the new node and becomes its
  category label.  This is a case of tree augmentation by node
  insertion.
\item If a syntactic constituent selected for grafting has already
  been labeled with a semantic tag, overlay the previous tag with the
  current tag.  We chose to tag in this manner simply because our
  system was not set up to handle the grafting of multiple tags onto a
  single constituent.  An example of this occurs in the sentence ``The
  Muslims had obtained Pakistan.''  If the NP node dominating {\it
  Pakistan} is grafted with a named entity tag such as NP-GPE, we
  overlay this with the NP-TargSucceed tag in a modality tagging
  scheme.
\item In the case of a word sequence, if the words covered by the
  semantic tag fall across two different syntactic constituents, do
  nothing.  This is a case of crossing brackets.
\end{itemize}

\begin{figure}
\begin{center}
\includegraphics[width=3.5in]{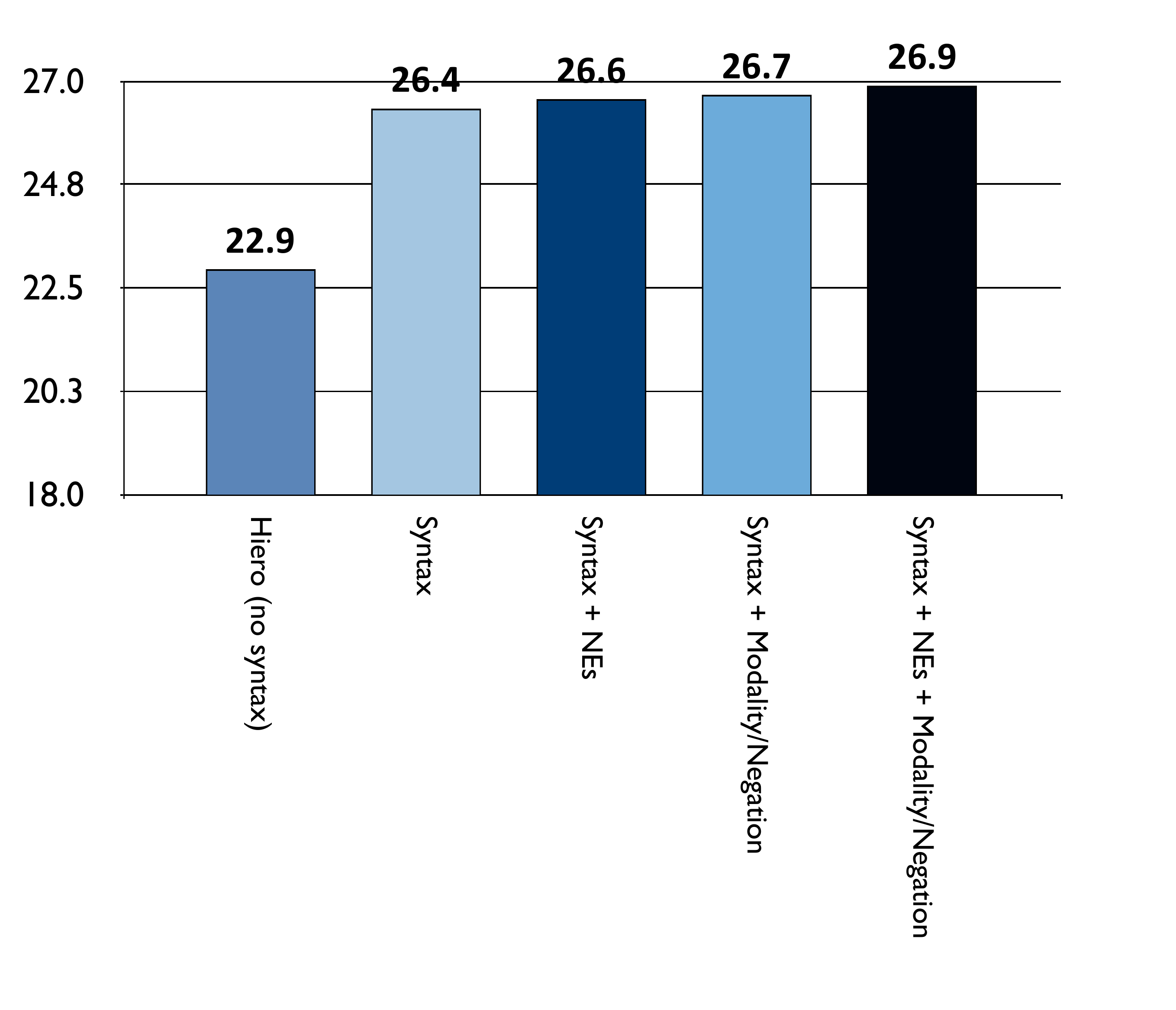}
\caption{Results for a range of experiments conducted during the SIMT effort show the score for our top-performing baseline systems derived from a hierarchical phrase-based model (Hiero). Substantial improvements obtained when syntax was introduced along with feature functions (FFs) and further improvements resulted from the addition of semantic elements. The scores are lowercased Bleu calculated on the held-out devtest set.}\label{joshua-experiments}
\end{center}
\end{figure}

Our tree-grafting procedure was simplified to accept a single semantic
tag per syntactic tree node as the final result.  The algorithm keeps
the last tag seen as the tag of precedence.  In practice, we
established a precedence ordering for modality/negation tags over
named entity tags by grafting named entity tags first and
modality/negation second.  Our intuition was that, in case of a tie,
finer-grained verbal categories would be more helpful to parsing than
finer-grained nominal categories.\footnote{In testing we found that
  grafting named entities first and modality/negation last yielded a
  slightly higher Bleu score than the reverse order.}  In cases where a word
was tagged both as a MN target and a MN
trigger, we gave precedence to the target tag.  This is because, while
MN targets vary, MN triggers are
generally identifiable with lexical items.  Finally, we used the
simplified specificity ordering of MN tags described in
Section~\ref{linguistic-simplifications} to ensure precedence of more
specific tags over more general ones. Table~\ref{Modalitytable} lists
the modality/negation types from highest (Require modality) to lowest (Negation) precedence.\footnote{Future work could include exploring
  additional methods of resolving tag conflicts or combining tag types
  on single nodes, e.g. by inserting multiple intermediate nodes
  (effectively using unary rewrite rules) or by stringing tag names
  together.}

\subsection{SIMT Results}
\label{results}

We evaluated our tree grafting approach by performing a series of translation experiments.  Each version of our translation system was trained on the same bilingual training data.  The bilingual parallel corpus that we used was distributed as part of the 2008 NIST Open Machine Translation Evaluation Workshop.\footnote{\url{http://www.itl.nist.gov/iad/mig/tests/mt/2008/doc/}}  The training set contained 88,108 Urdu-English sentence pairs, and a bilingual dictionary with 113,911 entries.  For our development and test sets, we split the NIST MT-08 test set into two portions (with each document going into either test or dev, and preserving the genre split).  Our test set contained 883 Urdu sentences, each with four translations into English, and our dev set contained 981 Urdu sentences, each with four reference translations.  To extract a syntactically informed translation model, we parsed the English side of the training corpus using a Penn Treebank trained parser~\cite{Miller:1998}.  For the experiments that involved grafting named entities onto the parse trees, we tagged the English side of the training corpus with the Phoenix tagger \cite{richman-schone:2008:ACLMain}.  We word-aligned the parallel corpus with the the Berkeley aligner.  All models used a 5-gram language model trained on the English Gigaword corpus (v5) using the SRILM toolkit with modified KN smoothing.  The Hiero translation grammar was extracted using the Joshua toolkit \cite{Li:2009}. The other translation grammars were extracted using the SAMT toolkit \cite{Venugopal2009}. 

Figure~\ref{joshua-experiments} gives the results for a number of
experiments conducted during the SIMT effort.\footnote{These
  experiments were conducted on the devtest set, containing 883 Urdu
  sentences (21,623 Urdu words) and four reference translations per
  sentence.  The Bleu score for these experiments is measured on
  uncased output.} The
experiments are broken into three groups: baselines, syntax, and
semantics.  To contextualize our results we experimented with a number
of different baselines that were composed from two different
approaches to statistical machine translation---phrase-based and
hierarchical phrase-based SMT---along with different combinations of
language model sizes and word aligners.  Our best-performing baseline
was a Hiero model.  The Bleu score for this baseline
on the development set was 22.9 Bleu points.

\begin{figure}
\begin{center}
\includegraphics[width=4.7in]{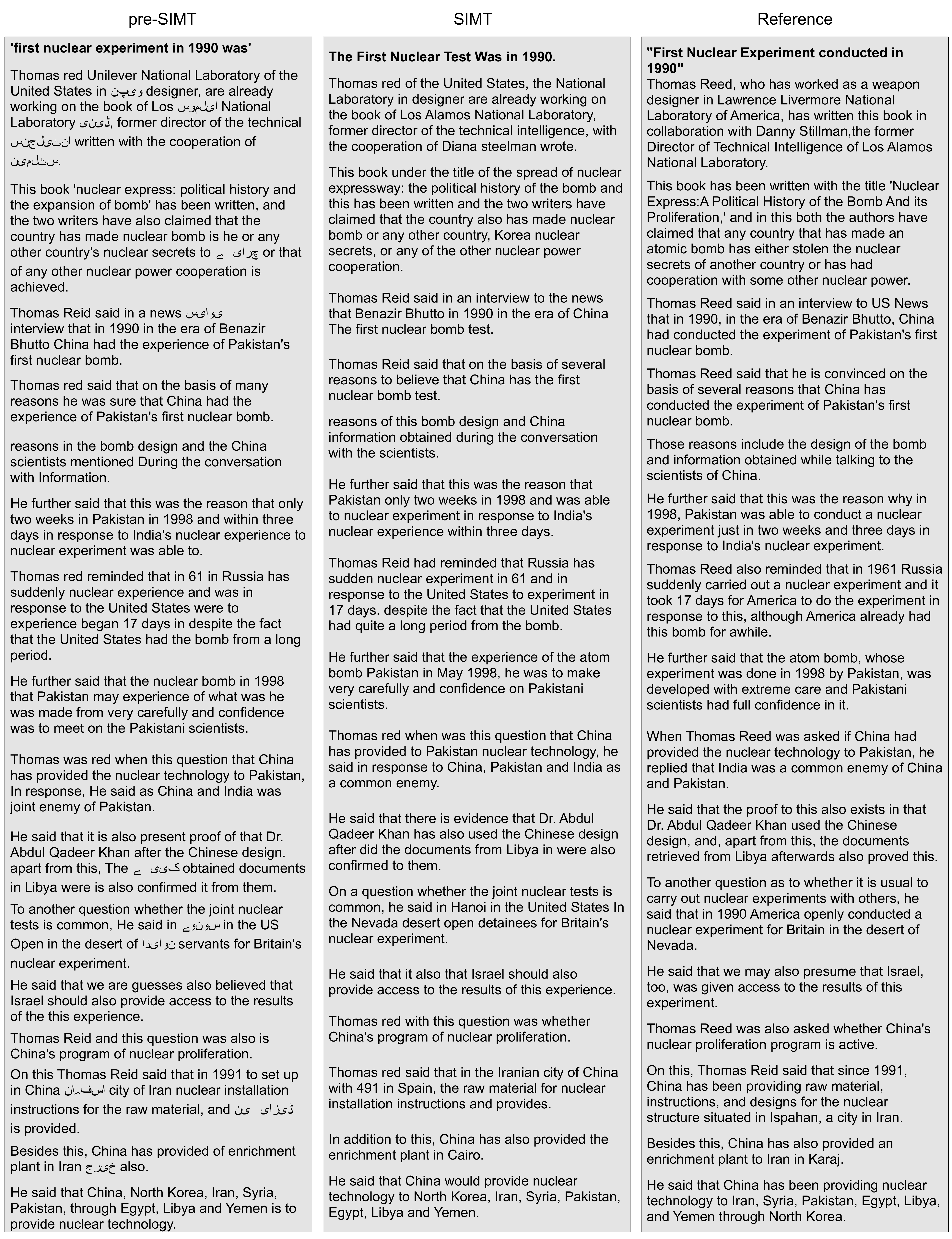}
\end{center}
\caption{An example of the improvements to Urdu-English translation before and after the SIMT effort. Output is from the baseline Hiero model, which does not use linguistic information, and from the final model, which incorporates syntactic and semantic information.  }
\label{final-output}
\end{figure}

After experimenting with syntactically motivated grammar rules, we
conducted 
experiments on the effects of incorporating semantic
elements (e.g., named entities and modality/negation) into the
translation grammars.  In our devtest set our taggers tagged on
average 3.5 named entities per sentence and 0.35 modality/negation
markers per
sentence.  These were included by grafting modality, negation, and 
named-entity markers
onto the parse trees.  Individually, each of these made modest
improvements over the syntactically-informed system alone.  Grafting
named entities onto the parse trees improved the Bleu score by 0.2
points.  Modality/negation improved it by 0.3 points.  Doing both
simultaneously had an additive effect and resulted in a 0.5 Bleu score
improvement over syntax alone.  This improvement was the largest
improvement that we got from anything other than the move from
linguistically na\"{\i}ve models to syntactically informed models.

We used bootstrap resampling to test whether the differences in Bleu scores were statistically significant \cite{koehn:2004:EMNLP}.  All of the results were a significant improvement over Hiero (at $p\leq0.01$).  The difference between the syntactic system and the syntactic system with NEs as not significant ( $p=0.38$). The differences between the  syntactic system and the syntactic system with MN, and between the  syntactic system and the syntactic system with both MN and NEs were both significant at ($p\leq0.05$).

Figure~\ref{final-output} shows example output from the final SIMT
system in comparison to the pre-SIMT results and the translation
produced by a human (reference).  An error analysis of this example
output illustrates that SIMT enhancements have resulted in the
elimination of misleading translation output in several cases:
\begin{enumerate}
\item 
{\bf pre-SIMT}: China had the experience of Pakistan's first nuclear bomb.\\
{\bf SIMT}: China has the first nuclear bomb test.\\
{\bf reference}: China has conducted the experiment of Pakistan's first nuclear bomb.
\item 
{\bf pre-SIMT}: the nuclear bomb in 1998 that Pakistan may experience\\
{\bf SIMT}: the experience of the atom bomb Pakistan in May 1998\\
{\bf reference}: the atom bomb, whose experiment was done in 1998 by Pakistan
\item 
{\bf pre-SIMT}: He said that it is also present proof of that Dr. Abdul Qadeer Khan after the Chinese design\\
{\bf SIMT}: He said that there is evidence that Dr. Abdul Qadeer Khan has also used the Chinese design\\
{\bf reference}: He said that the proof to this also exists in that Dr. Abdul Qadeer Khan used the Chinese design
\end{enumerate}
The article pertains to claims by Thomas Reid that China allowed
Pakistan to detonate a nuclear weapon at its test site.  However, in
the first example above, the reader is potentially misled by the
pre-SIMT output to believe that Pakistan launched a nuclear bomb on
China.  The SIMT output leaves out the mention of Pakistan, but
correctly conveys the firm belief that the bomb event is a {\it
  test\/} (closely resembling the term {\it experiment\/} in the human
reference), not a true bombing event. This is clearly an improvement
over the misleading pre-SIMT output.

In the second example, the pre-SIMT output misleads the reader to
believe that Pakistan is (or will be) attacked, through the use of the
phrase {\it may experience\/}, where {\it may\/} is poorly placed.
(We note here that this is a date translation error, i.e., the month
of {\it May\/} should be next to the year 1998, further adding to the
potential for confusion.)  Unfortunately, the SIMT output also uses
the term {\it experience\/} (rather than {\it experiment\/}, which is
in the human reference), but in this case the month is correctly
positioned in the output, thus eliminating the potential for confusion
with respect to the modality. The lack of a modal appropriately
neutralizes the statement so that it refers to an abstract event
associated with the atom bomb, rather than an attack on the country.

In the third example, where the Chinese design used by Dr. Abdul
Qandeer Khan is argued to be proof of the nuclear testing relationship
between Pakistan and China, the first pre-SIMT output potentially
leads the reader to believe that Dr. Abdul Qadeer is after the Chinese
design (not that he actually used it), whereas the SIMT output conveys
the firm belief that the Chinese design has been used by Dr. Abdul
Qadeer. This output very closely matches the human reference.

Note that even in the title of the article, the SIMT system produces
much more coherent English output than that of the linguistically
na\"{\i}ve system.  The figure also shows improvements due to
transliteration, which are described in~\cite{Irvine-PBML}.  The
scores reported in Figure~\ref{joshua-experiments} do not include
transliteration improvements.

\section{Conclusions and Future Work}
\label{conclusions}

We developed a modality/negation lexicon and a set of automatic MN taggers, one of
which---the structure-based tagger---results in 86\% precision for
tagging of a standard LDC data set. The MN tagger has been used
to improve machine translation output by imposing semantic constraints
on possible translations in the face of sparse training data.  The
tagger is also an important component of a language-understanding
module for a related project.

We have described a technique for translation that shows particular
promise for low-resource languages.  We have integrated linguistic
knowledge into statistical machine translation in a unified and
coherent framework.  We demonstrated that augmenting hierarchical
phrase-based translation rules with semantic labels (through
``grafting'') resulted in a 0.5 Bleu score improvement over syntax
alone.

Although our largest gains were from syntactic enrichments to the
Hiero model, demonstrating success on the integration of semantic
aspects of language bodes well for additional improvements based on the
incorporation of other semantic aspects. For example, we hypothesize that 
incorporating relations and temporal knowledge into the translation rules would further improve 
translation quality.  The syntactic grafting framework is well-suited to
support the exploration of the impact of many different types of
semantics on MT quality, though in this article we focused on exploring the impact of modality and negation.

An important future study is one that focuses on demonstrating whether
further improvements in modality/negation identification are likely to
lead to further gains in translation performance. Such a study would
benefit from the inclusion of a more detailed manual evaluation to
determine if modality and negation is adequately conveyed in the
downstream translations.  This work would be additionally enhanced
through experimentation on other language pair(s) and larger corpora.

The work presented here represents the first small steps toward a full
integration of MT and semantics.  Efforts underway in DARPA's GALE
program have already demonstrated the potential for combining MT and
semantics (termed {\it distillation\/}) to answer the information
needs of monolingual speakers using multilingual sources. 
Proper recognition of modalities and negation is crucial for handling those information needs effectively. 
In previous work, however, semantic processing proceeded largely independently of
the MT system, operating only on the translated output.  Our approach
is significantly different in that it combines syntax, semantics, and
MT into a single model, offering the potential advantages of joint
modeling and joint decision-making.  It would be interesting to
explore whether the integration of MT with syntax and semantics can be
extended to provide a single-model solution for tasks such as
cross-language information extraction and question answering, and to
evaluate our integrated approach, e.g., using GALE distillation
metrics.

\section*{Acknowledgments}

We thank Aaron Phillips for help with conversion of the output of the
entity tagger for ingest by the tree-grafting program.  We thank Anni
Irvine and David Zajic for their help with experiments on an
alternative Urdu modality/negation tagger based on projection and
training an HMM-based tagger derived from
Identifinder~\cite{Bikel:1999}.  For their helpful ideas and
suggestions during the development of the modality framework, we are
indebted to Mona Diab, Eduard Hovy, Marge McShane, Teruko Mitamura,
Sergei Nirenburg, Boyan Onyshkevych, and Owen Rambow.  We also thank
Basis Technology Corporation for their generous contribution of
software components to this work. This work was supported, in part, by
the Johns Hopkins Human Language Technology Center of Excellence (HLTCOE), by
the National Science Foundation under grant IIS-0713448, and by BBN
Technologies under GALE DARPA/IPTO Contract No. HR0011-06-C-0022. Any
opinions, findings, and conclusions or recommendations expressed in
this material are those of the authors and do not necessarily reflect
the views of the sponsor.

\bibliographystyle{fullname}
\bibliography{paper}

\begin{thebibliography}{}

\bibitem[\protect\citename{Baker, Fillmore, and Lowe}1998]{Baker:1998}
Baker, Collin~F., Charles~J. Fillmore, and John~B. Lowe.
\newblock 1998.
\newblock The {B}erkeley {F}rame{N}et project.
\newblock In {\em Proceedings of the 36th Annual Meeting of the Association for
  Computational Linguistics and 17th International Conference on Computational
  Linguistics - Volume 1}, ACL '98, pages 86--90, Stroudsburg, PA, USA.
  Association for Computational Linguistics.

\bibitem[\protect\citename{Baker \bgroup et al.\egroup }{2010}a]{Baker:2010a}
Baker, Kathryn, Steven Bethard, Michael Bloodgood, Ralf Brown, Chris
  Callison-Burch, Glen Coppersmith, Bonnie~J. Dorr, Nathaniel~W. Filardo,
  Kendall Giles, Ann Irvine, Michael Kayser, Lori Levin, Justin Martineau,
  James Mayfield, Scott Miller, Aaron Phillips, Andrew Philpot, Christine
  Piatko, Lane Schwartz, and David Zajic.
\newblock {2010}a.
\newblock {Semantically Informed Machine Translation}.
\newblock Technical Report {002}, {Johns Hopkins University, Baltimore, MD},
  {Human Language Technology Center of Excellence}.

\bibitem[\protect\citename{Baker \bgroup et al.\egroup }{2010}b]{Baker:2010d}
Baker, Kathryn, Michael Bloodgood, Chris Callison-Burch, Bonnie~J. Dorr,
  Nathaniel~W. Filardo, Lori Levin, Scott Miller, and Christine Piatko.
\newblock {2010}b.
\newblock Semantically-informed machine translation: A tree-grafting approach.
\newblock In {\em {Proceedings of The Ninth Biennial Conference of the
  Association for Machine Translation in the Americas}}, {Denver, CO}.

\bibitem[\protect\citename{Baker \bgroup et al.\egroup }{2010}c]{Baker:2010b}
Baker, Kathryn, Michael Bloodgood, Mona Diab, Bonnie~J. Dorr, Ed~Hovy, Lori
  Levin, Marjorie McShane, Teruko Mitamura, Sergei Nirenburg, Christine Piatko,
  Owen Rambow, and Gramm Richardson.
\newblock {2010}c.
\newblock {SIMT SCALE 2009 - Modality Annotation Guidelines}.
\newblock Technical Report {004}, {Johns Hopkins University, Baltimore, MD},
  {Human Language Technology Center of Excellence}.

\bibitem[\protect\citename{Baker \bgroup et al.\egroup }{2010}d]{Baker:2010c}
Baker, Kathryn, Michael Bloodgood, Bonnie~J. Dorr, Nathanial~W. Filardo, Lori
  Levin, and Christine Piatko.
\newblock {2010}d.
\newblock A modality lexicon and its use in automatic tagging.
\newblock In {\em {Proceedings of the Seventh International Conference on
  Language Resources and Evaluation (LREC)}}, pages {1402--1407},
  {Mediterranean Conference Center, Valletta, MALTA}.

\bibitem[\protect\citename{Bar-Haim \bgroup et al.\egroup }2007]{Bar-Haim:2007}
Bar-Haim, Roy, Ido Dagan, Iddo Greental, and Eyal Shnarch.
\newblock 2007.
\newblock Semantic inference at the lexical-syntactic level.
\newblock In {\em Proceedings of the 22nd National Conference on Artificial
  intelligence - Volume 1}, pages 871--876. AAAI Press.

\bibitem[\protect\citename{Bikel, Schwartz, and Weischedel}1999]{Bikel:1999}
Bikel, Daniel~M., Richard Schwartz, and Ralph~M. Weischedel.
\newblock 1999.
\newblock An algorithm that learns what's in a name.
\newblock {\em Machine Learning}, 34(1-3):211--231.

\bibitem[\protect\citename{B{\"{o}}hmov{\'a}, Cinkov{\'a}, and
  Haji{\v{c}}ov{\'a}}2005]{trmanEn2005}
B{\"{o}}hmov{\'a}, Alena, Silvie Cinkov{\'a}, and Eva Haji{\v{c}}ov{\'a}.
\newblock 2005.
\newblock {A Manual for Tectogrammatical Layer Annotation of the Prague
  Dependency Treebank (English translation)}.
\newblock Technical report, {\'U}FAL MFF UK, Prague, Czech Republic.

\bibitem[\protect\citename{Chiang}2005]{Chiang:2005}
Chiang, David.
\newblock 2005.
\newblock A hierarchical phrase-based model for statistical machine
  translation.
\newblock In {\em Proceedings of the 43rd Annual Meeting of the Association for
  Computational Linguistics (ACL-2005)}, Ann Arbor, Michigan.

\bibitem[\protect\citename{Diab \bgroup et al.\egroup }2009]{Diab:2009}
Diab, Mona~T., Lori Levin, Teruko Mitamura, Owen Rambow, Vinodkumar
  Prabhakaran, and Weiwei Guo.
\newblock 2009.
\newblock Committed belief annotation and tagging.
\newblock In {\em Proceedings of the Third Linguistic Annotation Workshop},
  ACL-IJCNLP '09, pages 68--73, Stroudsburg, PA, USA. Association for
  Computational Linguistics.

\bibitem[\protect\citename{Farkas \bgroup et al.\egroup }2010]{Farkas:2010}
Farkas, Rich\'{a}rd, Veronika Vincze, Gy\"{o}rgy M\'{o}ra, J\'{a}nos Csirik,
  and Gy\"{o}rgy Szarvas.
\newblock 2010.
\newblock The {CoNLL-2010} shared task: {L}earning to detect hedges and their
  scope in natural language text.
\newblock In {\em Proceedings of the Fourteenth Conference on Computational
  Natural Language Learning --- Shared Task}, CoNLL '10: Shared Task, pages
  1--12, Stroudsburg, PA, USA. Association for Computational Linguistics.

\bibitem[\protect\citename{Fellbaum}1998]{fellbaum98wordnet}
Fellbaum, Christiane, editor.
\newblock 1998.
\newblock {\em {WordNet: an electronic lexical database}}.
\newblock MIT Press.

\bibitem[\protect\citename{Haji{\v{c}} \bgroup et al.\egroup }2001]{PDT10}
Haji{\v{c}}, Jan, Eva Haji{\v{c}}ov{\'a}, Petr Pajas, Jarmila Panevov{\'a},
  Petr Sgall, and Barbora~Vidov{\'a} Hladk{\'a}.
\newblock 2001.
\newblock {Prague Dependency Treebank 1.0 (Final Production Label)}.

\bibitem[\protect\citename{Huang and Knight}2006]{HuangKnight06}
Huang, Bryant and Kevin Knight.
\newblock 2006.
\newblock Relabeling syntax trees to improve syntax-based machine translation
  quality.
\newblock In {\em HLT-NAACL}, New York.

\bibitem[\protect\citename{Irvine \bgroup et al.\egroup }2010]{Irvine-PBML}
Irvine, Ann, Mike Kayser, Zhifei Li, Wren Thornton, and Chris Callison-Burch.
\newblock 2010.
\newblock Integrating output from specialized modules in machine translation:
  Transliteration in {J}oshua.
\newblock {\em The Prague Bulletin of Mathematical Linguistics}, 93:107--116.

\bibitem[\protect\citename{Klein and Manning}2003]{KleinManning03}
Klein, Dan and Christopher~D. Manning.
\newblock 2003.
\newblock Accurate unlexicalized parsing.
\newblock In {\em Proceedings of the 41st Annual Meeting of the Association for
  Computational Linguistics}, pages 423--430.

\bibitem[\protect\citename{Koehn}2004]{koehn:2004:EMNLP}
Koehn, Philipp.
\newblock 2004.
\newblock Statistical significance tests for machine translation evaluation.
\newblock In Dekang Lin and Dekai Wu, editors, {\em Proceedings of EMNLP 2004},
  pages 388--395, Barcelona, Spain, July. Association for Computational
  Linguistics.

\bibitem[\protect\citename{Koehn \bgroup et al.\egroup }2007]{Moses}
Koehn, Philipp, Hieu Hoang, Alexandra Birch, Chris Callison-Burch, Marcello
  Federico, Nicola Bertoldi, Brooke Cowan, Wade Shen, Christine Moran, Richard
  Zens, Chris Dyer, Ondrej Bojar, Alexandra Constantin, and Evan Herbst.
\newblock 2007.
\newblock Moses: Open source toolkit for statistical machine translation.
\newblock In {\em Proceedings of the ACL-2007 Demo and Poster Sessions}.

\bibitem[\protect\citename{Kratzer}2009]{Kratzer}
Kratzer, Angelika.
\newblock 2009.
\newblock Plenary address at the annual meeting of the {L}inguistic {S}ociety
  of {A}merica.

\bibitem[\protect\citename{Larreya}2009]{Larreya:2009}
Larreya, Paul.
\newblock 2009.
\newblock Towards a typology of modality in language.
\newblock In Raphael Salkie, Pierre Busuttil, and Johan van~der Auwera,
  editors, {\em Modality in English: {T}heory and description}. Mouton de
  Gruyter, pages 9--30.

\bibitem[\protect\citename{Li \bgroup et al.\egroup }2009]{Li:2009}
Li, Zhifei, Chris Callison-Burch, Chris Dyer, Sanjeev Khudanpur, Lane Schwartz,
  Wren Thornton, Jonathan Weese, and Omar Zaidan.
\newblock 2009.
\newblock {Joshua}: An open source toolkit for parsing-based machine
  translation.
\newblock In {\em Proceedings of the Fourth Workshop on Statistical Machine
  Translation}, pages 135--139, Athens, Greece, March. Association for
  Computational Linguistics.

\bibitem[\protect\citename{Marcus, Marcinkiewicz, and
  Santorini}1993]{Marcus:93}
Marcus, Mitchell~P., Mary~Ann Marcinkiewicz, and Beatrice Santorini.
\newblock 1993.
\newblock Building a large annotated corpus of {E}nglish: the {P}enn
  {T}reebank.
\newblock {\em Computational Linguistics}, 19(2):313--330.

\bibitem[\protect\citename{McShane, Nirenburg, and
  Zacharsky}2004]{McShaneEtAl:2004}
McShane, Marjorie, Sergei Nirenburg, and Ron Zacharsky.
\newblock 2004.
\newblock Mood and modality: Out of the theory and into the fray.
\newblock {\em Natural Language Engineering}, 19(1):57--89.

\bibitem[\protect\citename{Miller \bgroup et al.\egroup }1998]{Miller:1998}
Miller, Scott, Heidi Fox, Lance Ramshaw, and Ralph Weischedel.
\newblock 1998.
\newblock {SIFT}: Statistically-derived information from text.
\newblock In {\em Seventh {M}essage {U}nderstanding {C}onference ({MUC}-7)},
  Washington, D.C.

\bibitem[\protect\citename{Miller \bgroup et al.\egroup }2000]{Miller:2000}
Miller, Scott, Heidi~J. Fox, Lance~A. Ramshaw, and Ralph~M. Weischedel.
\newblock 2000.
\newblock A novel use of statistical parsing to extract information from text.
\newblock In {\em Proceedings of Applied Natural Language Processing and the
  North American Association for Computational Linguistics}.

\bibitem[\protect\citename{Murata \bgroup et al.\egroup }2005]{Murata:2005}
Murata, Masaki, Kiyotaka Uchimoto, Qing Ma, Toshiyuki Kanamaru, and Hitoshi
  Isahara.
\newblock 2005.
\newblock Analysis of machine translation systems' errors in tense, aspect, and
  modality.
\newblock In {\em Proceedings of the 19th Asia-Pacific Conference on Language,
  Information and Computation}.

\bibitem[\protect\citename{Nairn, Condorovdi, and Karttunen}2006]{Nairn:2006}
Nairn, Rowan, Cleo Condorovdi, and Lauri Karttunen.
\newblock 2006.
\newblock Computing relative polarity for textual inference.
\newblock In {\em Proceedings of the International Workshop on Inference in
  Computational Semantics}, ICoS-5.

\bibitem[\protect\citename{Nirenburg and McShane}2008]{NirenburgMcShane}
Nirenburg, Sergei and Marjorie McShane.
\newblock 2008.
\newblock {The formulation of modalities (speaker attitude) in {O}nto{S}em}.

\bibitem[\protect\citename{Palmer, Gildea, and Kingsbury}2005]{Palmer:2005}
Palmer, Martha, Daniel Gildea, and Paul Kingsbury.
\newblock 2005.
\newblock The {P}roposition {B}ank: An annotated corpus of semantic roles.
\newblock {\em Computational Linguistics}, 31:71--106, March.

\bibitem[\protect\citename{Papineni \bgroup et al.\egroup }2002]{Papineni:2002}
Papineni, Kishore, Salim Roukos, Todd Ward, and Wei-Jing Zhu.
\newblock 2002.
\newblock Bleu: A method for automatic evaluation of machine translation.
\newblock In {\em Proceedings of the 40th Annual Meeting of the Association for
  Computational Linguistics (ACL-2002)}, Philadelphia, Pennsylvania.

\bibitem[\protect\citename{Petrov and Klein}2007]{Petrov-Klein-2007:AAAI}
Petrov, Slav and Dan Klein.
\newblock 2007.
\newblock Learning and inference for hierarchically split {PCFG}s.
\newblock In {\em AAAI 2007 (Nectar Track)}.

\bibitem[\protect\citename{Prabhakaran, Rambow, and
  Diab}2010]{Prabhakaran:2010}
Prabhakaran, Vinodkumar, Owen Rambow, and Mona Diab.
\newblock 2010.
\newblock Automatic committed belief tagging.
\newblock In {\em Proceedings of the 23rd International Conference on
  Computational Linguistics: Posters}, COLING '10, pages 1014--1022,
  Stroudsburg, PA, USA. Association for Computational Linguistics.

\bibitem[\protect\citename{Prasad \bgroup et al.\egroup }2008]{PRASAD08.754}
Prasad, Rashmi, Nikhil Dinesh, Alan Lee, Eleni Miltsakaki, Livio Robaldo,
  Aravind Joshi, and Bonnie Webber.
\newblock 2008.
\newblock The {P}enn {D}iscourse {T}ree{B}ank 2.0.
\newblock In Nicoletta Calzolari~(Conference Chair), Khalid Choukri, Bente
  Maegaard, Joseph Mariani, Jan Odjik, Stelios Piperidis, and Daniel Tapias,
  editors, {\em Proceedings of the Sixth International Language Resources and
  Evaluation (LREC'08)}, Marrakech, Morocco, May. European Language Resources
  Association (ELRA).

\bibitem[\protect\citename{Pustejovsky \bgroup et al.\egroup
  }2006]{Pustejovsky06}
Pustejovsky, James, Marc Verhagen, Roser Saur\'{i}, Jessica Littman, Robert
  Gaizauskas, Graham Katz, Inderjeet Mani, Robert Knippen, and Andrea Setzer.
\newblock 2006.
\newblock {\em TimeBank 1.2}.
\newblock Linguistic Data Consortium, Philadelphia.

\bibitem[\protect\citename{Richman and
  Schone}2008]{richman-schone:2008:ACLMain}
Richman, Alexander and Patrick Schone.
\newblock 2008.
\newblock Mining wiki resources for multilingual named entity recognition.
\newblock In {\em Proceedings of ACL-08: HLT}, pages 1--9, Columbus, Ohio,
  June. Association for Computational Linguistics.

\bibitem[\protect\citename{Rubin}2007]{Rubin07}
Rubin, Victoria~L.
\newblock 2007.
\newblock Stating with certainty or stating with doubt: Intercoder reliability
  results for manual annotation of epistemically modalized statements.
\newblock In {\em HLT-NAACL (Short Papers)}, pages 141--144.

\bibitem[\protect\citename{Saur\'{i} and Pustejovsky}2009]{Sauri09}
Saur\'{i}, Roser and James Pustejovsky.
\newblock 2009.
\newblock {F}act{B}ank: A corpus annotated with event factuality.
\newblock {\em Language Resources and Evaluation}, 43(3):227--268.

\bibitem[\protect\citename{Saur\'{i}, Verhagen, and
  Pustejovsky}2006]{SauriVP06}
Saur\'{i}, Roser, Marc Verhagen, and James Pustejovsky.
\newblock 2006.
\newblock Annotating and recognizing event modality in text.
\newblock In Geoff Sutcliffe and Randy Goebel, editors, {\em FLAIRS
  Conference}, pages 333--339. AAAI Press.

\bibitem[\protect\citename{Sigurd and Gawr\'{o}nska}1994]{Sigurd:1994}
Sigurd, Bengt and Barbara Gawr\'{o}nska.
\newblock 1994.
\newblock Modals as a problem for {MT}.
\newblock In {\em Proceedings of the 15th conference on Computational
  linguistics - Volume 1}, COLING '94, pages 120--124, Stroudsburg, PA, USA.
  Association for Computational Linguistics.

\bibitem[\protect\citename{Steedman}1999]{Steedman:1999}
Steedman, Mark.
\newblock 1999.
\newblock Alternating quantifier scope in {CCG}.
\newblock In {\em Proceedings of the 37th Annual Meeting of the Association for
  Computational Linguistics (ACL)}, College Park, Maryland.

\bibitem[\protect\citename{Szarvas \bgroup et al.\egroup }2008]{Szarvas:2008}
Szarvas, Gy\"{o}rgy, Veronika Vincze, Rich\'{a}rd Farkas, and J\'{a}nos Csirik.
\newblock 2008.
\newblock The {BioScope} corpus: {A}nnotation for negation, uncertainty and
  their scope in biomedical texts.
\newblock In {\em Proceedings of the Workshop on Current Trends in Biomedical
  Natural Language Processing}, BioNLP '08, pages 38--45, Stroudsburg, PA, USA.
  Association for Computational Linguistics.

\bibitem[\protect\citename{van~der Auwera and Ammann}2005]{VanDerAuweraAmman}
van~der Auwera, Johan and Andreas Ammann.
\newblock 2005.
\newblock Overlap between situational and epistemic modal marking.
\newblock In Martin Haspelmath, Matthew~S. Dryer, David Gil, and Bernard
  Comrie, editors, {\em World Atlas of Language Structures}. Oxford University
  Press, chapter~76, pages 310--313.

\bibitem[\protect\citename{Venugopal and Zollmann}2009]{Venugopal2009}
Venugopal, Ashish and Andreas Zollmann.
\newblock 2009.
\newblock Grammar based statistical {MT} on {Hadoop}: An end-to-end toolkit for
  large scale {PSCFG} based {MT}.
\newblock {\em Prague Bulletin of Mathematical Linguistics}, 91.

\bibitem[\protect\citename{Venugopal, Zollmann, and Vogel}2007]{Venugopal:2007}
Venugopal, Ashish, Andreas Zollmann, and Stephan Vogel.
\newblock 2007.
\newblock An efficient two-pass approach to synchronous-{CFG} driven
  statistical {MT}.
\newblock In {\em Proceedings of the Human Language Technology Conference of
  the North American chapter of the Association for Computational Linguistics
  (HLT/NAACL-2007)}, Rochester, New York.

\bibitem[\protect\citename{von Fintel and Iatridou}2009]{VonFintelIatridou}
von Fintel, Kai and Sabine Iatridou.
\newblock 2009.
\newblock Morphology, syntax, and semantics of modals.
\newblock Lecture notes for 2009 {LSA} {I}nstitute class.

\bibitem[\protect\citename{Wang \bgroup et al.\egroup }2010]{WangEtAl:2010}
Wang, Wei, Jonathan May, Kevin Knight, and Daniel Marcu.
\newblock 2010.
\newblock Re-structuring, re-labeling, and re-aligning for syntax-based machine
  translation.
\newblock {\em Computational Linguistics}, 36(2).

\bibitem[\protect\citename{Webber \bgroup et al.\egroup }2003]{Webber:2003}
Webber, Bonnie, Aravid Joshi, Matthew Stone, and Alistair Knott.
\newblock 2003.
\newblock Anaphora and discourse structure.
\newblock {\em Computational Linguistics}, 29:545--587.

\bibitem[\protect\citename{Wiebe, Wilson, and Cardie}2005]{Wiebe05}
Wiebe, Janyce, Theresa Wilson, and Claire Cardie.
\newblock 2005.
\newblock Annotating expressions of opinions and emotions in language.
\newblock {\em Language Resources and Evaluation}, 39(2-3):165--210.

\bibitem[\protect\citename{Wilson, Wiebe, and Hoffman}2009]{Wilson:2009}
Wilson, Theresa, Janyce Wiebe, and Paul Hoffman.
\newblock 2009.
\newblock Recognizing contextual polarity: An exploration of features for
  phrase-level sentiment analysis.
\newblock {\em Computational Linguistics}, 35:399--433, September.

\bibitem[\protect\citename{Zollmann and Venugopal}2006]{ZollmannVenugopal:2006}
Zollmann, Andreas and Ashish Venugopal.
\newblock 2006.
\newblock Syntax augmented machine translation via chart parsing.
\newblock In {\em Proceedings on the Workshop on Statistical Machine
  Translation}, pages 138--141, New York City, June. Association for
  Computational Linguistics.

\end{thebibliography}

\end{document}